\documentclass[journal]{IEEEtran}
\usepackage[square, comma, sort&compress, numbers]{natbib}
\usepackage{amsmath,amsfonts}
\usepackage{algorithmic}
\usepackage{array}
\usepackage[caption=false,font=scriptsize,labelfont=rm,textfont=rm, subrefformat=parens]{subfig}

\usepackage[colorlinks,urlcolor=blue,linkcolor=blue,citecolor=blue]{hyperref}
\usepackage{color,array,xcolor}
\usepackage{graphicx}
\usepackage{textcomp}
\usepackage{stfloats}
\usepackage{url}
\usepackage{bm}
\usepackage{makecell}
\usepackage{verbatim}
\usepackage{graphicx}
\usepackage{hyperref}
\usepackage{booktabs}
\usepackage{multirow}
\hypersetup{breaklinks=true}
\def\BibTeX{{\rm B\kern-.05em{\sc i\kern-.025em b}\kern-.08em
    T\kern-.1667em\lower.7ex\hbox{E}\kern-.125emX}}

\usepackage{balance}
\begin{document}
\title{StreakNet-Arch: An Anti-scattering Network-based Architecture for Underwater Carrier LiDAR-Radar Imaging}

\author{
  Xuelong~Li$^{\dagger}$,~\IEEEmembership{Fellow, IEEE},
  Hongjun~An$^{\dagger}$, 
  Haofei~Zhao,
  Guangying~Li,
  Bo~Liu,
  Xing~Wang,
  Guanghua~Cheng,
  Guojun~Wu,
  and~Zhe~Sun$^{\ast}$
\thanks{$^{\dagger}$Xuelong~Li and Hongjun~An contributed equally to this work.}
\thanks{$^{\ast}$Corresponding author: Zhe~Sun (sunzhe@nwpu.edu.cn).}
\thanks{This research was supported by the China National Key R\&D Program (2022YFC2808003), the Fundamental Research Funds for the Central Universities (D5000220481), and the Natural Science Foundation of Shaanxi Province, P. R. China (2024JC-YBMS-468).}
\thanks{Xuelong~Li, Hongjun~An, Haofei~Zhao, Guanghua~Cheng and Zhe~Sun are with School of Artificial Intelligence, OPtics and ElectroNics (iOPEN), Northwestern Polytechnical University, Xi'an 710072, Shaanxi, P. R. China. Xuelong~Li, Hongjun~An, Haofei~Zhao and Zhe~Sun are also with the Institute of Artificial Intelligence (TeleAI), China Telecom, Shanghai 200000, P. R. China.}
\thanks{Guangying~Li is with the State Key Laboratory of Transient Optics and Photonics, Xi'an Institute of Optics and Precision Mechanics of CAS, Xi'an 710119, Shaanxi, P. R. China.}
\thanks{Bo~Liu and Guojun~Wu are with the Marine Optical Technology Laboratory, Xi'an Institute of Optics and Precision Mechanics of CAS, Xi'an 710119, Shaanxi, P. R. China.}
\thanks{Xing~Wang is with the Key Laboratory of Spectral Imaging Technology, Xi'an Institute of Optics and Precision Mechanics of CAS, Xi'an 710119, Shaanxi, P. R. China.}}

\markboth{IEEE Transactions on Image Processing,~Vol.~xx, No.~xx, ~xxxx}%
{How to Use the IEEEtran \LaTeX \ Templates}

\maketitle

\begin{abstract}

In this paper, we introduce StreakNet-Arch, a real-time, end-to-end binary-classification framework based on our self-developed Underwater Carrier LiDAR-Radar (UCLR) that embeds Self-Attention and our novel Double Branch Cross Attention (DBC-Attention) to enhance scatter suppression. Under controlled water tank validation conditions, StreakNet-Arch with Self-Attention or DBC-Attention outperforms traditional bandpass filtering and achieves higher $F_1$ scores than learning-based MP networks and CNNs at comparable model size and complexity. Real-time benchmarks on an NVIDIA RTX 3060 show a constant Average Imaging Time (54 to 84 ms) regardless of frame count, versus a linear increase (58 to 1,257 ms) for conventional methods. To facilitate further research, we contribute a publicly available streak-tube camera image dataset contains 2,695,168 real-world underwater 3D point cloud data. More importantly, we validate our UCLR system in a South China Sea trial, reaching an error of 46mm for 3D target at 1,000 m depth and 20 m range. Source code and data are available at \href{https://github.com/BestAnHongjun/StreakNet}{https://github.com/BestAnHongjun/StreakNet}.
\end{abstract}

\begin{IEEEkeywords}
Underwater laser imaging, Signal processing, Streak-tube camera, LiDAR-Radar, Attention mechanism.
\end{IEEEkeywords}

\section{Introduction}

\IEEEPARstart{U}{nderwater} laser imaging signal processing technology is crucial for obtaining underwater images, including 2D gray-scale maps and 3D point clouds images, which has wide applications in ocean exploration, biology \cite{9416821}, surveillance \cite{8607026}, archaeology, unmanned underwater vehicles control \cite{10045793, 9262026}, etc. In contrast to image processing algorithms for underwater image enhancement \cite{10129222, 10155564, 9930878, 10048777, 9855418, 9854113, 9832540, 9426457, 10382428} or restoration \cite{10238432, 7840002, 9541354}, underwater laser imaging signal processing technology can process signals from a more fundamental source, such as streak-tube camera and ICCD camera. This approach enables the achievement of superior spatial resolution and extended detection ranges. However, its effectiveness is significantly hindered by a major challenge: scattering. This phenomenon drastically reduces image clarity and limits imaging range.

To address this, the Underwater Carrier LiDAR-Radar (UCLR) employs a suite of strategies to suppress scattering and achieve long-distance underwater imaging \cite{mullen1994modulated, mullen1995microwave, mullen2000hybrid, 10.1117/12.2050395}. Specifically, the UCLR's laser source typically utilizes blue or green light to minimize propagation attenuation in water \cite{sun2025water, cariou1982transmission}, thereby enhancing detection distance. Additionally, a range-gated detector is employed for the UCLR, which is sensitive only to reflected signals received within a specific time window after the pulse is emitted. More importantly, lasers are modulated into high-frequency pulses to exceed the cut-off frequency of water's low-pass response \cite{pellen2000determination,pellen_radio_2001}, effectively suppressing light scattering. Since the frequency is typically high ($\geq$100 MHz), receivers employing high temporal resolution optical detection devices are required, such as nanosecond-resolution ICCD camera \cite{takahashi2020observation} or picosecond-resolution streak-tube camera \cite{li2021lidar, fang2024streak, fang2022development}.
Underwater laser imaging relies on signal processing algorithms to extract target echoes from the received signal. These algorithms determine the presence and arrival time of the echoes, ultimately reconstructing the image. The processing typically involves two stages: scatter suppression and echo identification. Scatter suppression methods in the UCLR include bandpass filtering \cite{li2021lidar}, adaptive filtering \cite{1170536, 1163096, 1170926, 1141414, 1163432, 1163949}, and machine learning-based filtering \cite{illig2020machine}. The objective is to process a signal containing scatter noise into a suppressed scatter signal. Echo identification methods in UCLR primarily rely on thresholding techniques, including manually set thresholds  \cite{li2021lidar} and adaptive thresholds \cite{otsu1975threshold}, often coupled with matched filtering approaches \cite{li2021lidar}. These methods aim to determine the presence of echo signals in the received signal.

However, despite demonstrably mitigating scattering effects, these algorithms exhibit limitations in two key areas. Considering one aspect, low filtering accuracy leads to the loss of valuable information within the signal processing. Bandpass filtering algorithms rely on manually designed filters \cite{li2021lidar}, where the bandpass range is determined empirically by engineers and may not necessarily be optimal. Alternatively, limitations in either algorithm complexity or real-time performance hinder their use for real-time underwater laser imaging. Adaptive filtering algorithms were primarily explored from the 1960s to the 1980s \cite{1170536, 1163096, 1170926, 1141414, 1163432, 1163949}, and the existing machine learning filtering algorithms \cite{illig2020machine} mainly rely on traditional McCulloch-Pitts (MP) neural networks \cite{Chakraverty2019}. Constrained by the computational capabilities of hardware available at that time, these models have a limited number of parameters, resulting in a relatively low upper limit on performance. Moreover, the current two-stage signal processing paradigm fails to achieve real-time imaging. This limitation arises from the echo identification in the second stage. Here, determining the threshold for identifying echoes requires denoising all collected scene signals and analyzing their statistical amplitude characteristics \cite{li2021lidar, otsu1975threshold}. This limitation severely constrains the practical utility of the UCLR.

\begin{figure*}
  \label{fig:overview}
  \centering
  \includegraphics[width=\textwidth]{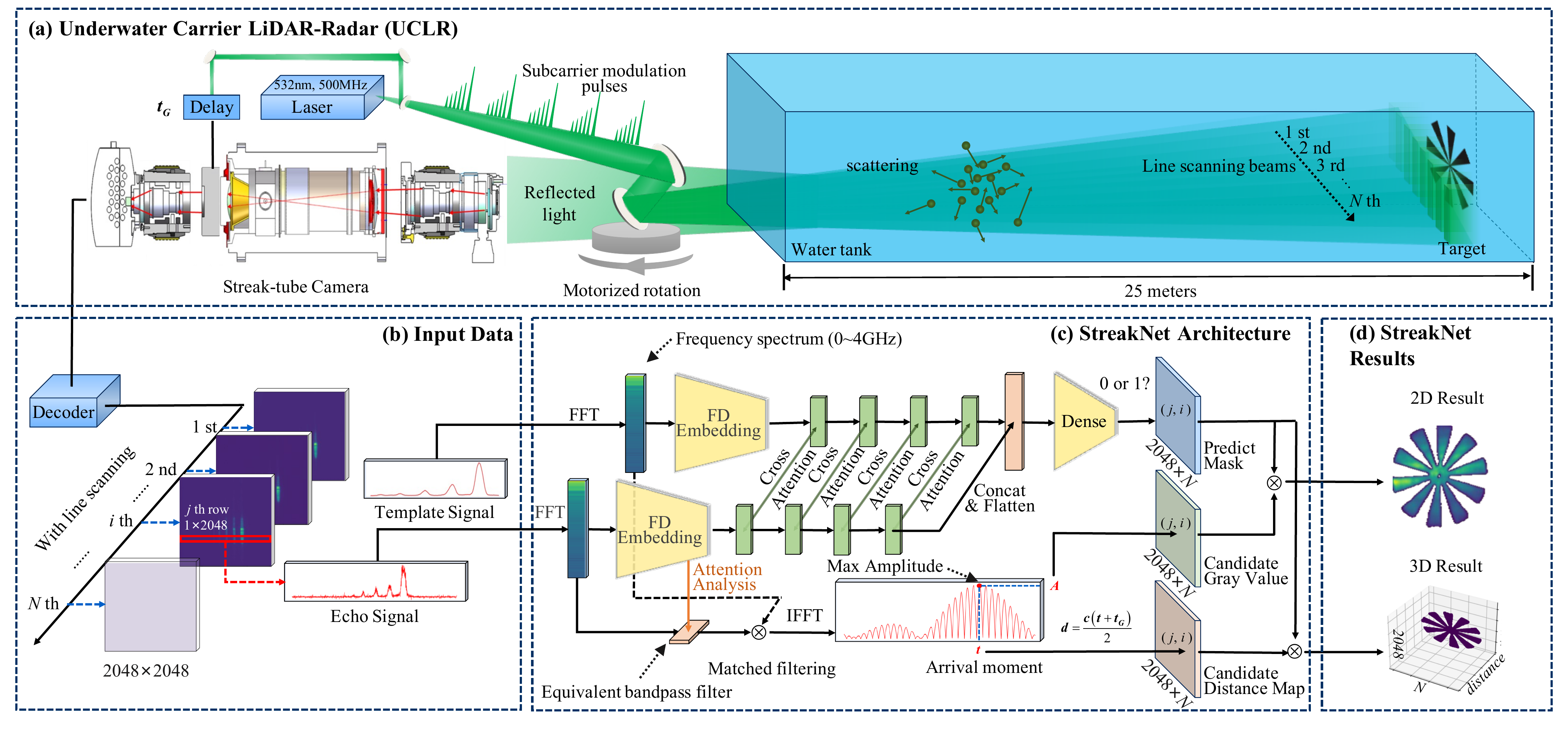}
  \caption{Overview of the StreakNet-Arch based UCLR system. (a) Pulses from a sub-nanosecond 500 MHz Q-switch laser (532 nm) are split for $t_G$-delayed streak-tube triggering and for scanning water-tank targets at $N$ discrete angles via a motorized turntable. (b) Decoding yields $N$ streak-tube images ($2048\times2048$), one per angle, where the horizontal axis represents time and the vertical axis represents space. Row $j$ in image $i$ encodes the 30 ns light-intensity echo signal at position $j$ and angle $i$. (c) Given the echo signal and template signal, the StreakNet-Arch-based UCLR system outputs (d) the $(j,i)$ pixel in both the $2048\times N$ grayscale and depth maps.}
\end{figure*}

In this paper, we firstly experimented with employing Self-Attention mechanism networks \cite{vaswani2017attention} in the signal processing phase of the self-developed UCLR to improve scatter-resistance. This architecture already SOTA in computer vision \cite{hu2018squeeze, wang2018non, guo2021pct, dosovitskiy2020image, yuan2021tokens} and NLP \cite{vaswani2017attention, devlin2018bert, yang2019xlnet}, emerges as a powerful universal model. To prevent overfitting and boost generalization across scenes, we provide a template signal alongside each input, guiding the network to learn echo-vs-noise distinctions. We further adapt Self-Attention into a Double Branch Cross Attention (DBC-Attention) mechanism, which our validation experiments show yields higher $F_1$ scores than both traditional bandpass filtering and contemporaneous learning-based MP and CNN methods at comparable model size and complexity (Table \ref{tab_f1_streaknet}) under controlled water tank environment.

Moreover, by recasting imaging as an end-to-end binary classification, our StreakNet-Arch directly flags echo-containing frames, eliminating the batch-wide pending time of conventional algorithms. On an NVIDIA RTX 3060 GPU, StreakNet-Arch achieves a constant Average Imaging Time (AIT) of 54 to 84 ms across up to 64 frames, whereas traditional methods’ AIT grows linearly from 58 ms to 1,257 ms (Fig. \ref{fig:ait_benchmark}, Table \ref{tab_ait_benchmark}), confirming its real-time advantage.

Given that our input comes from streak-tube camera captures, we name this end-to-end framework StreakNet-Arch. Finally, to validate deep-sea performance, we conducted a South China Sea trial, reaching an error of 46mm for 3D target at 1,000 m depth and 20 m range.

The main contributions of this paper can be summarized as follows: 

\begin{enumerate}{}{}
  \item{We introduce StreakNet-Arch, a novel end-to-end binary classification architecture that revolutionizes the UCLR's signal processing. This approach empowers the UCLR with real-time imaging capabilities for the first time.}
  \item{We enhance the UCLR's signal processing with Self-Attention networks. Further, we propose DBC-Attention, a groundbreaking variant specifically optimized for underwater imaging tasks. Experimental results under controlled water tank environment conclusively demonstrate DBC-Attention's superiority over the standard Self-Attention approach.}
  \item{We propose a method to embed streak-tube camera images directly into the attention network. This embedded representation effectively functions as a learned bandpass filter, as demonstrated by our experiments.}
  \item{We released a large-scale dataset containing 2,695,168 real-world underwater 3D point cloud data captured by streak-tube camera, which facilitates further development of Underwater laser imaging signal processing techniques.}
  \item{We validated the UCLR system in a deep-sea field experiment in the South China Sea, reaching an error of 46mm for 3D target at 1,000 m depth and 20 m.}
\end{enumerate}

\section{Related Work}

\subsection{Signal processing algorithms of UCLR}

The signal processing algorithms for underwater laser imaging can be broadly categorized into two stages: scatter suppression and echo identification. Scatter suppression aims to process a signal containing scatter noise into a scatter-suppressed signal. Conventional methods primarily involve bandpass filtering \cite{li2021lidar}, where engineers define a frequency bandpass range based on their experiential knowledge to suppress clutter noise. However, this approach is limited by the subjective expertise of engineers and may not always yield optimal results. From the 1960s to the 1980s,  researchers explored various adaptive filtering techniques to address limitations in bandpass filtering. These techniques, including lattice filters \cite{1170536} and least squares lattice algorithms \cite{1163096}, operate in the time domain.  Additionally, there were frequency domain methods such as the LMS algorithm \cite{1141414} and its variants like FLMS \cite{1163432} and UFLMS \cite{1163949}. Subsequently, scholars combined machine learning algorithms based on MP neural networks to achieve adaptive clutter suppression \cite{illig2020machine}.

In the UCLR, echo identification methods primarily rely on thresholding techniques. These encompass manually setting thresholds \cite{li2021lidar} and adaptive thresholding \cite{otsu1975threshold}, often in conjunction with matched filtering methodologies \cite{li2021lidar}, with the aim of identifying the presence of echo signals within the input signal.

\subsection{Attention Mechanism}

In the past decade, the attention mechanism has played an increasingly important role in computer vision and natural language processing. In 2014, Mnih V. et al. \cite{mnih2014recurrent} pioneered the use of attention mechanism into neural networks, predicting crucial regions through policy gradient recursion and updating the entire network end-to-end. Subsequent works \cite{xu2015show, gregor2015draw} in visual attention leveraged recurrent neural networks (RNNs) as essential tools. Hu J. et al. proposed SENet \cite{hu2018squeeze}, presenting a novel channel-attention network that implicitly and adaptively predicts potential key features. A significant shift came in 2017 with the introduction of the Self-Attention mechanism by Vaswani et al \cite{vaswani2017attention}. This advancement revolutionized Natural Language Processing (NLP) \cite{devlin2018bert, yang2019xlnet}. In 2018, Wang et al. \cite{wang2018non} took the lead in introducing Self-Attention to computer vision. Notably, Hu et al. (2018) proposed a channel-attention network (SENet) within this timeframe. Recently, various Self-Attention networks (Visual Transformers, ViTs) \cite{guo2021pct, dosovitskiy2020image, yuan2021tokens, zhuang2022gsam, zhai2022lit} have appeared, showcasing the immense potential of attention-based models.

Attention mechanisms can also be applied to the enhancement of underwater image processing \cite{10129222, ummar2023window, 10445832}. In 2023, Peng L. et al. introduced the U-shape Transformer, pioneering the incorporation of self-attention mechanisms into underwater image enhancement \cite{10129222}. They proposed a Transformer module that fuses multi-scale features across channels, and a spatial module for global feature modeling. This innovation enhances the network's focus on areas of more severe attenuation in both color channels and spatial regions. Mehnaz U. et al. proposed an innovative Underwater window-based Transformer Generative Adversarial Network (UwTGAN) aimed at enhancing underwater image quality for computer vision applications in marine settings \cite{ummar2023window}. Pramanick A. at el. propose a framework that considered wavelength of light in underwater conditions by using cross-attention transformers \cite{10445832}.

\section{Method}

\subsection{StreakNet-Arch} \label{subsec:uclr}

The proposed StreakNet-Arch based self-developed UCLR system (Fig. \hyperref[fig:overview]{1a-d}) employs a sub-nanosecond Q-switch laser to generate subcarrier-modulated pulses at a frequency of 500 MHz with 532 nm, 80 mJ. A portion of the generated pulse passes through the beam splitter into a delay device, which can be gated by a delay of $t_G$ seconds. \footnote{Range-gated imaging technology, which captures images by controlling the camera shutter delay for a certain period $t_G$. This enables the reception of signals within a specific range, mitigating the impact of backscattering on imaging.} After the delay, a trigger signal is sent to the control circuit of the streak-tube camera. Simultaneously, another part of the pulse is reflected into the water tank.

By rotating the motorized turntable, a line scan of remote underwater objects is achieved. The reflected light from the objects reaches the streak-tube camera. Upon decoding, a series of streak-tube images is generated (Fig. \hyperref[fig:overview]{1b}).

For the line scan containing $N$ discrete angles, the system will generate $N$ streak-tube images, each with dimensions of $2048\times 2048$. The horizontal axis corresponds to the full-screen scanning time at that angle, while the vertical axis corresponds to space. For the $j$-th row of the $i$-th image, it represents the $j$-th ($0 \leq j < 2048$) vertical spatial position for the $i$-th scanning angle ($0 \leq i < N$), with the light intensity variation over 30ns time sampled as a $1\times 2048$ vector. 

After inputting this vector along with a corresponding template signal vector into the StreakNet-Arch based UCLR system, the resulting output will correspond to the $(j,i)$ component of both a 2D grayscale map and a 3D depth map, where the dimensions of both maps are 2048$\times N$
(Fig. \hyperref[fig:overview]{1d}). 

\subsection{FD Embedding Layer}

In section \ref{subsec:uclr}, we introduce that the StreakNet-Arch's inputs consist of an echo timing signal vector $\mathbf{v}_{echo} \in \mathbb{R}^{1\times N_s}$ and a template timing signal vector $\mathbf{v}_{tem} \in \mathbb{R}^{1\times L_{tem}}(L_{tem} \leq N_s)$, where $N_s=2048$ in our project. With a full-screen scan time of $T_{\text{full}}=30$ ns and $N_s$ samples taken, the signal vector is sampled at a frequency of 68.27 GHz (see Eq. \ref{eq_sampling}).

\begin{equation}
  \label{eq_sampling}
  f_s = \frac{N_s}{T_{\text{full}}} , \Delta R_{f} = \frac{fs}{N_{\text{FFT}}} .
\end{equation}

The two vectors will be firstly fed into the Frequency Domain (FD) Embedding Layer (FDEL) of the network. Upon entering the FDEL, the vectors will undergo a Fast Fourier Transform (FFT). During the transformation, the lengths of the two vectors will be standardized by padding with zeros up to $N_{\text{FFT}}$,  to obtain an appropriate frequency resolution $\Delta R_{f}$ after the transformation. In our work, we set $N_{\text{FFT}}$ to be $2^{16}$, hence the frequency resolution is approximately 1 MHz. (see Eq. \ref{eq_sampling}).

% \begin{equation}
%   \label{eq_resolution}
%   \Delta R_{f} = \frac{fs}{N_{\text{FFT}}} .
% \end{equation}

After the transformation, a spectrum of length $N_{\text{FFT}}$ will be obtained, corresponding to a frequency range of $0$ to $f_s / 2$. However, the carrier frequency $f_c$ is typically much smaller than $f_s / 2$, so only the portion of the frequency vector from index $0$ to $L=k\lceil f_c / \Delta R_f \rceil$ is usually retained, where $k$ is a correction factor. (see Eq. \ref{eq_fft}). In our work, we set $L$ to be $4000$, meaning only frequency components up to approximately 4 GHz are retained.

\begin{equation}
  \label{eq_fft}
  \begin{split}
    \mathbf{u}_{\text{echo}} &= \mathcal{FFT}(\mathbf{v}_{\text{echo}}, N_{\text{FFT}})[0:L] ,\\
    \mathbf{u}_{\text{tem}} &= \mathcal{FFT}(\mathbf{v}_{\text{tem}}, N_{\text{FFT}})[0:L] .
  \end{split}
\end{equation}

It is worth noting that from an engineering point of view, the current neural network under the PyTorch framework \cite{imambi2021pytorch} does not support vector inputs of imaginary numbers. So we introduce an imaginary expansion operator (IEO) (see Eq. \ref{eq_ieo}) to convert the imaginary vector ($\mathbf{u} \in \mathbb{C}^{1 \times L}$) to a real vector ($\mathbf{u'} \in \mathbb{R}^{1 \times 2L}$).

\begin{equation}
  \label{eq_ieo}
  \text{IEO:}
  \mathbf{u'}_{k} = 
  \begin{cases}
    \mathbf{Re}(\mathbf{u}_k), & 0 \leq k < L, \\
    \mathbf{Im}(\mathbf{u}_{k-L}), & L \leq k < 2L .
  \end{cases}
\end{equation}

\begin{equation}
  \label{eq_apply_ieo}
    \mathbf{u'}_{\text{echo}} = \text{IEO}(\mathbf{u}_{\text{echo}}) , \mathbf{u'}_{\text{tem}} = \text{IEO}(\mathbf{u}_{\text{tem}}) .
\end{equation}

After applying IEO (see Eq. \ref{eq_apply_ieo}), two vectors of length $2L$ are obtained. Clearly, not every component significantly contributes to the recognition task. Therefore, a linear layer (see Eq. \ref{eq_embedding_linear}) is subsequently applied for feature extraction. Now introducing a width factor, denoted as $\lambda_{w}$ ($0 \leq \lambda_{w} \leq 1$, with official recommendations of 0.125, 0.25, 0.50, or 1.00), the input dimension of the linear layer is set to $2L$, and the output dimension is $\lfloor 512\lambda_{w} \rfloor$.

\begin{equation}
  \label{eq_embedding_linear}
  \begin{split}
    \mathbf{X}_{\text{echo}}^\top &= \text{SiLU}(\mathbf{W}_{\text{echo}} \mathbf{u'}_{\text{echo}}^{\top} + b_{\text{echo}}) ,\\
    \mathbf{X}_{\text{tem}}^\top &= \text{SiLU}(\mathbf{W}_{\text{tem}} \mathbf{u'}_{\text{tem}}^{\top} + b_{\text{tem}}) ,
  \end{split}
\end{equation}

\noindent where $\mathbf{W}_* \in \mathbb{R}^{\lfloor 512\lambda_{w} \rfloor \times 2L}$ and $b_{*} \in \mathbb{R}$ are respectively the learnable weight matrix and bias. The $\mathbf{X}_* \in \mathbb{R}^{1 \times \lfloor 512\lambda_{w} \rfloor}$ are outputs of the FDEL.

\subsection{Attention Analysis Method}

We leverage attention analysis to elucidate the learning mechanism of the FDEL. Our experiments demonstrate that the FDEL effectively functions as a learned bandpass filter.

From the perspective of MP neuron model \cite{Chakraverty2019}, the linear layer is essentially a series of input nodes and MP neurons, and the weight matrix is the connection weight between input nodes and neurons. If we want to calculate the input of $j$-th neuron, we need to multiply all the input nodes by their respective weights and then sum them (see Eq. \ref{eq_mp_model}).

\begin{equation}
  \label{eq_mp_model}
  \text{y}_{j} = \sum_{i=0}^{N_{\text{input}}}{w_{ij} \cdot x_i} .
\end{equation}

\begin{figure}[ht]
  \centering
  \subfloat[]{
    \includegraphics[width=.4\columnwidth]{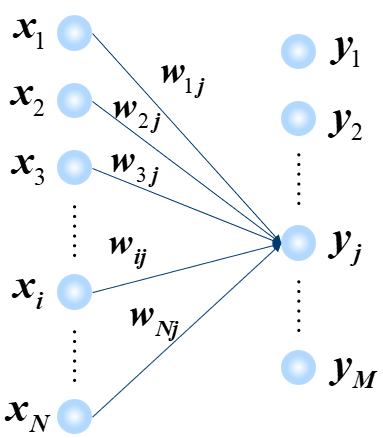}
    \label{fig:view_nn}
  }\hspace{8pt}
  \subfloat[]{
    \includegraphics[width=.4\columnwidth]{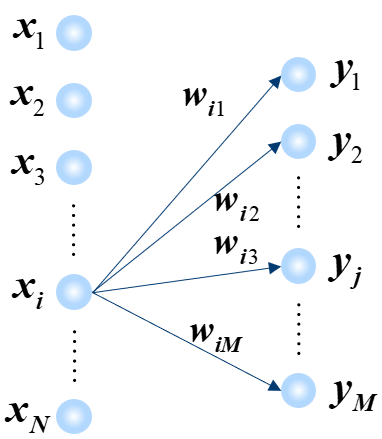}
    \label{fig:view_input}
  }
  \caption{Different perspectives of MP model. (a) From the neuron’s perspective, the input to neuron $j$ is computed by summing the products of all input nodes and their associated weights. (b) From the input node’s perspective, if input node $i$ connects to neuron $j$ with weight $w_{ij}$, then neuron $j$ extracts information proportional to $\Vert w_{ij} \Vert$ from node $i$, indicating the attention allocated by neuron $j$ to node $i$.}
  \label{fig:view}
\end{figure}

This perspective is from the viewpoint of neurons (Fig. \ref{fig:view_nn}). If we reverse the view to consider it from the perspective of input nodes (Fig. \ref{fig:view_input}), for input node $i$, if there is a connection weight $w_{ij}$ with neuron $j$, it implies that neuron $j$ has extracted the quantity of information $\Vert w_{ij} \Vert$ from node $i$. In other words, neuron $j$ has allocated its attention to node $i$ through the weight $\Vert w_{ij} \Vert$. If neuron $j$ is completely indifferent to the information from node $i$, then $w_{ij}$ should be equal to 0. In that case, the total attention of the neural network to input node $i$ should be expressed by Eq. \ref{eq_attention_analysis}.

\begin{equation}
  \label{eq_attention_analysis}
  {A'}_{i} = \sum_{j=0}^{N_{\text{neurons}}}{\Vert w_{ij} \Vert} .
\end{equation}

Standardize the attention to unify units (see Eq. \ref{eq_attention}).The above process can be called Attention Analysis Method (AAM). If we perform an Attention Analysis on the weight matrix $\mathbf{W}_{\text{echo}}$ of the FDEL, the resulting attention distribution can be equivalent to the transfer function of a bandpass filter.

\begin{equation}
  \label{eq_attention}
  {A}_{i} = \frac{{A'}_{i}-\min\{{A'}_{i}\}}{\max\{{A'}_{i}\}-\min\{{A'}_{i}\}} , \mathbf{A} = \left(A_i\right)_{1\times 2L},
\end{equation}

\noindent where $\mathbf{A}$ is the filtering transfer function.

\subsection{Double Branch Cross Attention Backbone}

Double Branch Cross Attention (DBC-Attention) is a special attention mechanism. For the input of two branches $\mathbf{X}_{\text{echo}},\mathbf{X}_{\text{tem}} \in \mathbb{R}^{1 \times \lfloor 512\lambda_{w} \rfloor}$, they are alternatively utilized as keys, values, and queries to compute the attention scores. Subsequently, upon aggregating the attention, the double branch deep feature tensors $\mathbf{Y}_{\text{echo}},\mathbf{Y}_{\text{tem}} \in \mathbb{R}^{1 \times \lfloor 512\lambda_{w} \rfloor}$ are generated through a nonlinear feedforward network.

The formal representation is as follows: Firstly, the keys, values, and queries are computed (see Eq. \ref{eq_dbca}).

\begin{equation}
  \label{eq_dbca}
  \begin{split}
    \mathbf{Q}_1 = \mathbf{W}_{q1}\mathbf{X}_{\text{echo}},\mathbf{Q}_2 = \mathbf{W}_{q2}\mathbf{X}_{\text{tem}} ,\\ 
    \mathbf{K}_1 = \mathbf{W}_{k1}\mathbf{X}_{\text{tem}},\mathbf{K}_2 = \mathbf{W}_{k2}\mathbf{X}_{\text{echo}} ,\\ 
    \mathbf{V}_1 = \mathbf{W}_{v1}\mathbf{X}_{\text{tem}},\mathbf{V}_2 = \mathbf{W}_{v2}\mathbf{X}_{\text{echo}} ,
  \end{split}
\end{equation}

\noindent where $\mathbf{W}_*$ are learnable parameters. Then, attention scores are computed and attention is aggregated. The residual method is employed by adding it to the input and followed by Layer Normalization (LNorm) \cite{ba2016layer} (see Eq. \ref{eq_agg_attention}):

\begin{equation}
  \label{eq_agg_attention}
  \begin{split}
    \mathbf{Y}_1 = \text{LNorm}\left[\mathbf{X}_{\text{echo}}+\text{softmax}\left(\frac{\mathbf{Q}_1\mathbf{K}_1^\top}{\sqrt{d_k}}\right)\mathbf{V}_1\right] ,\\ 
    \mathbf{Y}_2 = \text{LNorm}\left[\mathbf{X}_{\text{tem}}+\text{softmax}\left(\frac{\mathbf{Q}_2\mathbf{K}_2^\top}{\sqrt{d_k}}\right)\mathbf{V}_2\right] ,
  \end{split}
\end{equation}

\noindent where $d_k$ is the column space dimension of the input/output tensor. Finally, the deep feature tensor is output through the feedforward layer. The residual method is also used here (see Eq. \ref{eq_ffl}).

\begin{equation}
  \label{eq_ffl}
  \begin{split}
    \mathbf{Y}_{\text{echo}} = \text{SiLU}\left[\text{LNorm}\left(\mathbf{W}_1\mathbf{Y}_1^\top+\mathbf{Y}_1^\top+b_1\right)\right] ,\\
    \mathbf{Y}_{\text{tem}} = \text{SiLU}\left[\text{LNorm}\left(\mathbf{W}_2\mathbf{Y}_2^\top+\mathbf{Y}_2^\top+b_2\right)\right] ,
  \end{split}
\end{equation}

\noindent where $\mathbf{W}_*$ and $b_*$ are learnable parameters, and \text{SiLU} \cite{DBLP:journals/corr/abs-1710-05941} is a type of nonlinear activation function.

Eq. \ref{eq_dbca}-\ref{eq_ffl} together form the basic block of DBC-Attention. Similar to the Transformer architecture \cite{vaswani2017attention}, DBC-Attention can use a multi-head attention approach when calculating scores. 

By stacking different numbers of DBC-Attention blocks, we can obtain backbone networks with different depths for DBC-Attention architecture.

\subsection{Imaging Head}

The Imaging Head comprises two data paths: denoising and imaging. The denoising path, modeled as a binary classification task, identifies target regions within the input feature tensor using a learned mask map, replacing traditional hand-crafted thresholds. The imaging path leverages traditional methods but incorporates a learned filter (replacing handcrafted bandpass filters) obtained through AAM during filtering. This results in candidate gray and distance maps. Finally, element-wise multiplication of the denoising mask with these maps generates the final imaging outputs.

\begin{itemize}
  \item[$\bullet$] \bf{Denoising path:}
\end{itemize}

The output $\mathbf{Y}_{\text{echo}},\mathbf{Y}_{\text{tem}}$ from backbone network is concatenated, and then passed through a feedforward layer to obtain a binary probability vector $\mathbf{Y}$ (see Eq. \ref{eq_denoise_path}).

\begin{equation}
  \label{eq_denoise_path}
    \mathbf{Y}^\top = \text{SiLU}\left(\mathbf{W}\cdot\text{Concat}\left(\mathbf{Y}_{\text{echo}},\mathbf{Y}_\text{tem}\right)^\top+b\right) ,
\end{equation}

\noindent where $\mathbf{W}$ and $b$ are learnable parameters. The mask map $\mathbf{M}$ is calculated using Eq. \ref{eq_mask_map}:

\begin{equation}
  \label{eq_mask_map}
  \mathbf{M}(j, i) = \text{argmax}(\mathbf{Y}) .
\end{equation}

\begin{itemize}
  \item[$\bullet$] \bf{Imaging path:}
\end{itemize}

First, the vector obtained from Eq. \ref{eq_ieo} is multiplied by the transfer function obtained through the AAM method (Eq. \ref{eq_attention}) to perform filtering operations (see Eq. \ref{eq_filter}).

\begin{equation}
  \label{eq_filter}
    \bm{\mu}'_{\text{echo}}=\mathbf{u}'_{\text{echo}} \odot \mathbf{A}.
\end{equation}

Next, the Inverse Imaginary Expansion Operator (IIEO) (Eq. \ref{eq_iieo}) is used to transform the real vector $\bm{\mu}'_{\text{echo}}\in \mathbb{R}^{1 \times 2L}$ into a complex vector $\bm{\mu}_{\text{echo}} \in \mathbb{C}^{1 \times L}$ (see Eq. \ref{eq_iieo_trans}).

\begin{equation}
  \label{eq_iieo}
  \text{IIEO}(\bm{\mu}')\text{:}\bm{\mu}_k=\bm{\mu}'_k+i\bm{\mu}'_{k+L} .
\end{equation}

\begin{equation}
  \label{eq_iieo_trans}
  \bm{\mu}_{\text{echo}} = \text{IIEO}(\bm{\mu}'_{\text{echo}}) .
\end{equation}

Then, multiply $\bm{\mu}_{\text{echo}}$ by the spectrum of template signal $\mathbf{u}_{\text{tem}}$, perform frequency-domain matched filtering, and transform back to the time domain using inverse fast Fourier transform (Eq. \ref{eq_ifft}).

\begin{equation}
  \label{eq_ifft}
  \mathbf{v}_f=\mathcal{IFFT}(\bm{\mu}_{\text{echo}} \odot \mathbf{u}_{\text{tem}}, N_{\text{FFT}})[0:N_s] ,
\end{equation}

\noindent where $\mathbf{v}_f \in \mathbb{R}^{1 \times N_s}$ time domain signal of scattering suppression. The candidate gray map ($\mathbf{CG}$) and candidate distance map ($\mathbf{CD}$) can be calculated as follows (Eq. \ref{eq_cal_map}):

\begin{equation}
  \label{eq_cal_map}
  \begin{split}
    i = \text{argmax}(\mathbf{v}_f) , & t = i\frac{1}{f_s} + t_G , \\
    \mathbf{CG}(j,i) =\max (\mathbf{v}_f) , & \mathbf{CD}(j,i) = \frac{c}{n} \cdot \frac{t}{2} ,
  \end{split}
\end{equation}

\noindent where $f_s$ is the sample frequency (Eq. \ref{eq_sampling}), $t_G$ is the gate time, $c$ the speed of light in vacuum, and $n$ is the refractive index of the propagation medium.

\begin{itemize}
  \item[$\bullet$] \bf{Path aggregation:}
\end{itemize}

By multiplying with the mask map $\mathbf{M}$, we obtain the gray map $\mathbf{G}$ and distance map $\mathbf{D}$ (see Eq. \ref{eq_image}).

\begin{equation}
  \label{eq_image}
    \mathbf{G} = \mathbf{CG} \odot \mathbf{M} , \mathbf{D} = \mathbf{CD} \odot \mathbf{M} .
\end{equation}

\subsection{Loss Function}

The loss function is the objective optimization function during the training phase. It is worth noting that although the Imaging Head contains a denoising path and imaging path, only the denoising path participates in the training process. The echo signal vector $\mathbf{v}_{\text{echo}}$ and the template signal vector $\mathbf{v}_{\text{tem}}$ sequentially pass through FD Embedding Layer, backbone network, and the denoising path of the Imaging Head to obtain a binary probability vector $\mathbf{Y}$, which represents the complete forward propagation process. Since this task can be modeled as a binary classification task, we choose cross-entropy as the loss function (Eq. \ref{eq_loss}).

\begin{equation}
  \label{eq_loss}
  \mathcal{L}(\mathbf{Y}, \mathbf{Y}') = -\sum_{i=0}^{2} Y'_i \log(Y_i) .
\end{equation}

\begin{figure*}[htbp]
  \centering
  \subfloat[]{
    \includegraphics[width=0.6\columnwidth]{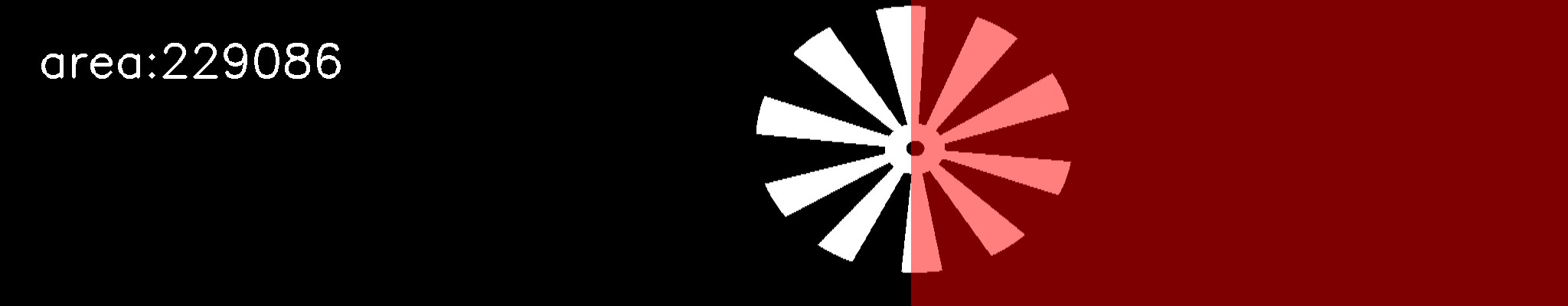}
    \label{fig:dataset_20m_train}
  }
  \subfloat[]{
    \includegraphics[width=0.6\columnwidth]{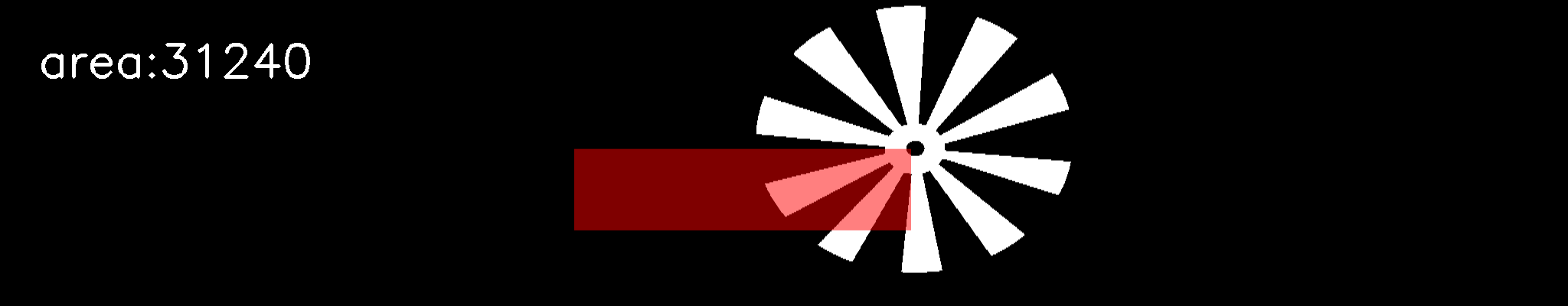}
    \label{fig:dataset_20m_valid}
  }
  \subfloat[]{
    \includegraphics[width=0.6\columnwidth]{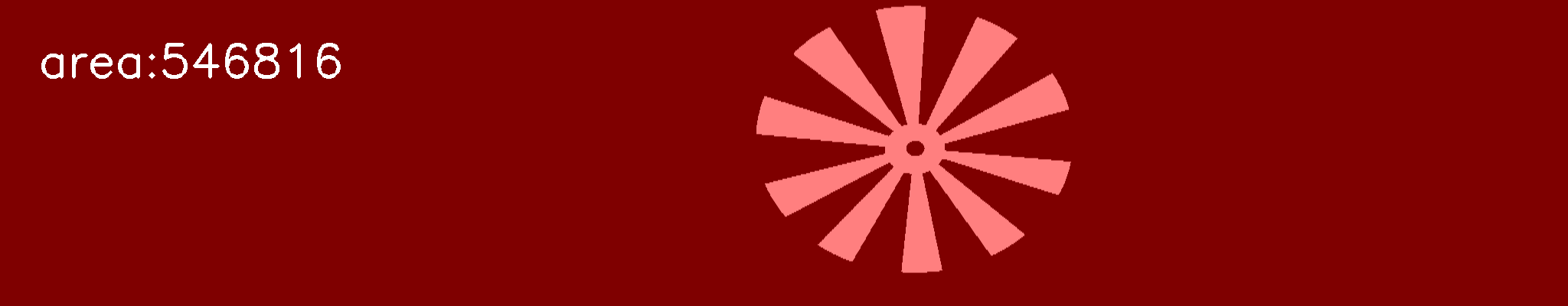}
    \label{fig:dataset_20m_test}
  } \\

  \subfloat[]{
    \includegraphics[width=0.6\columnwidth]{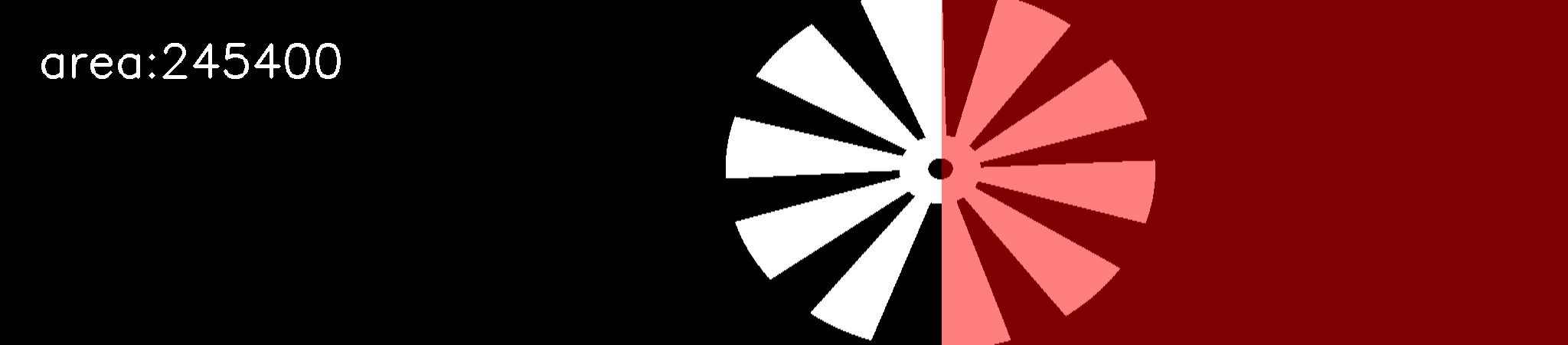}
  }
  \subfloat[]{
    \includegraphics[width=0.6\columnwidth]{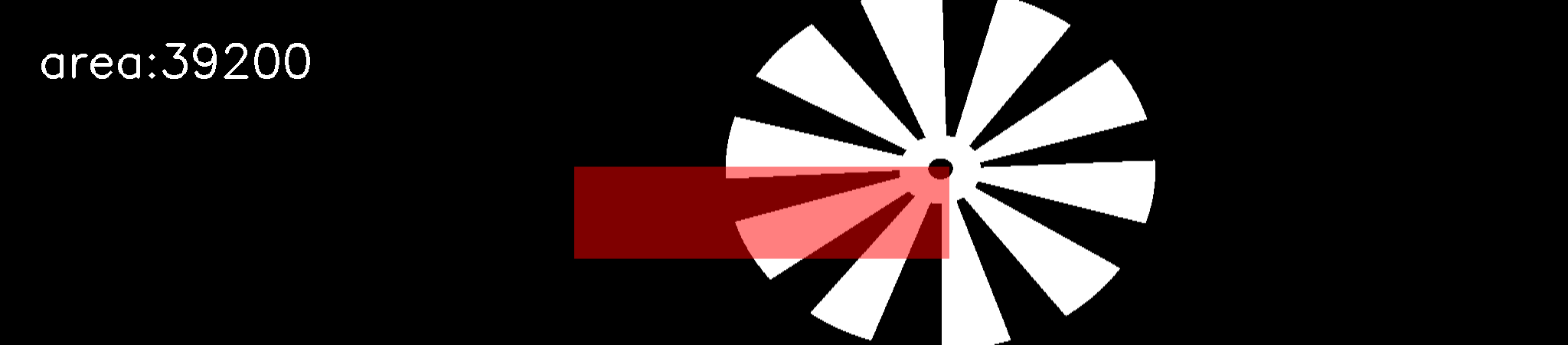}
  }
  \subfloat[]{
    \includegraphics[width=0.6\columnwidth]{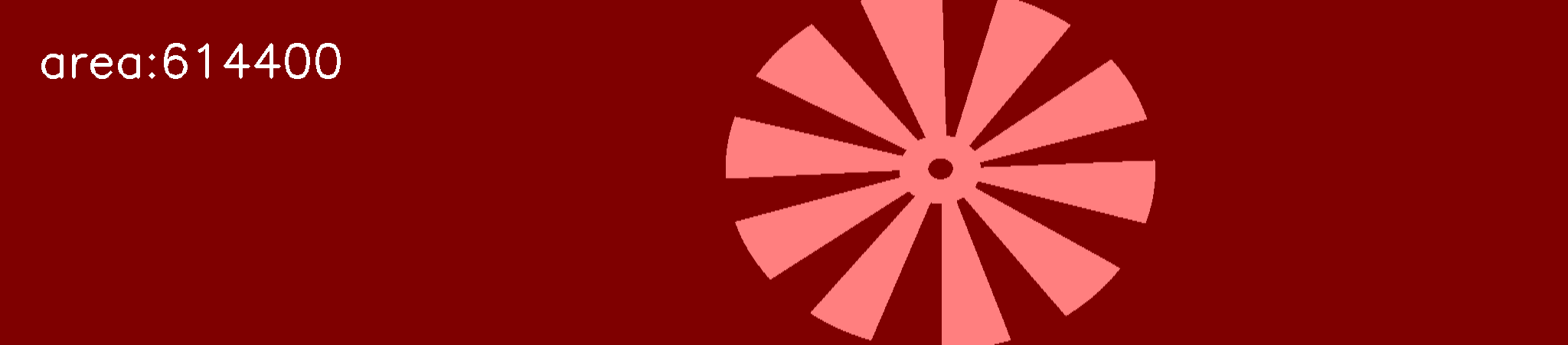}
  } \\

  \subfloat[]{
    \includegraphics[width=0.6\columnwidth]{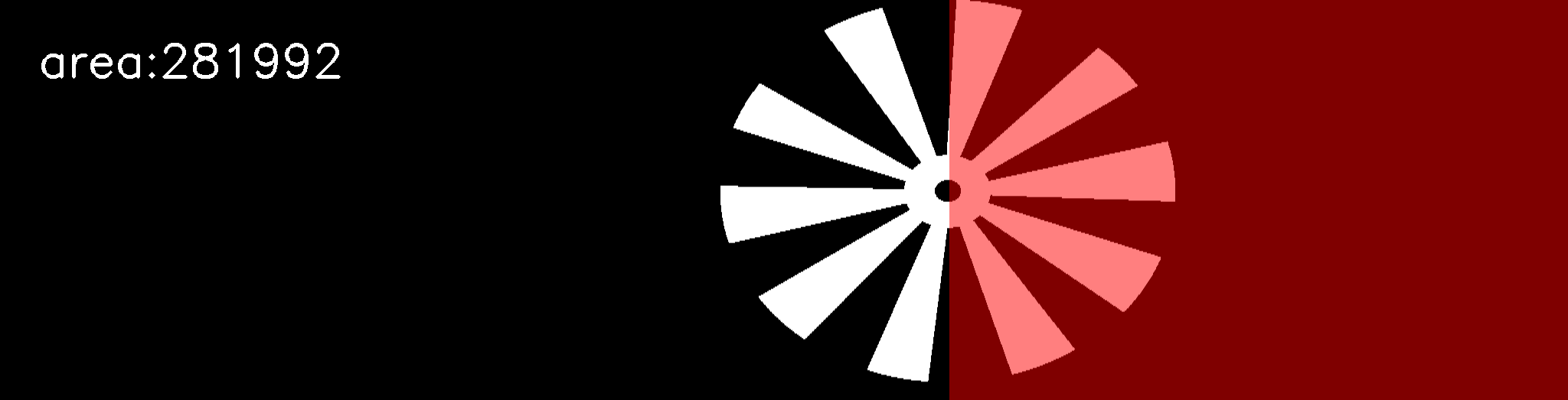}
  }
  \subfloat[]{
    \includegraphics[width=0.6\columnwidth]{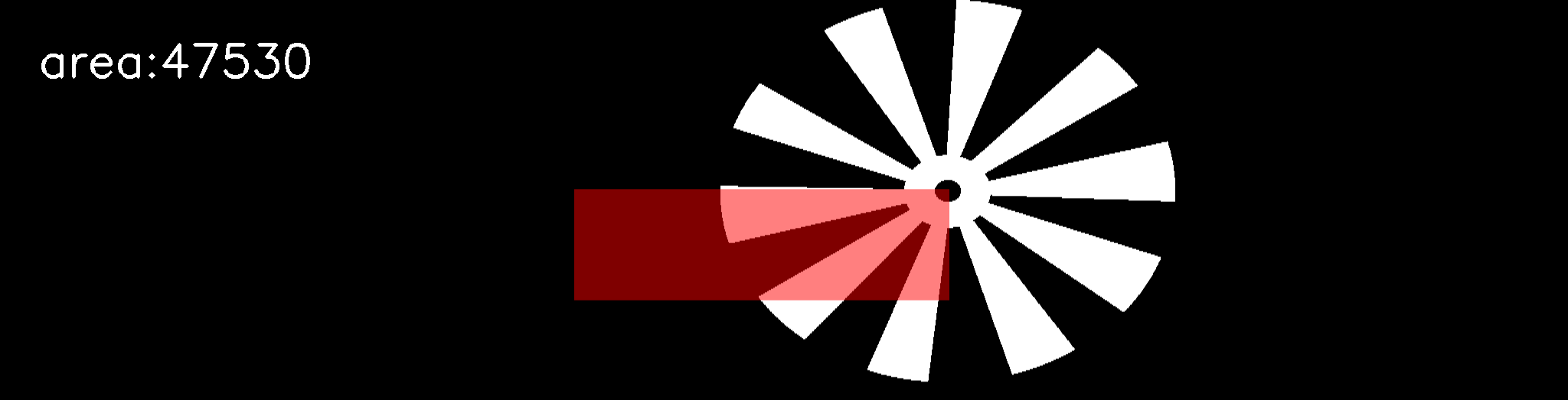}
  }
  \subfloat[]{
    \includegraphics[width=0.6\columnwidth]{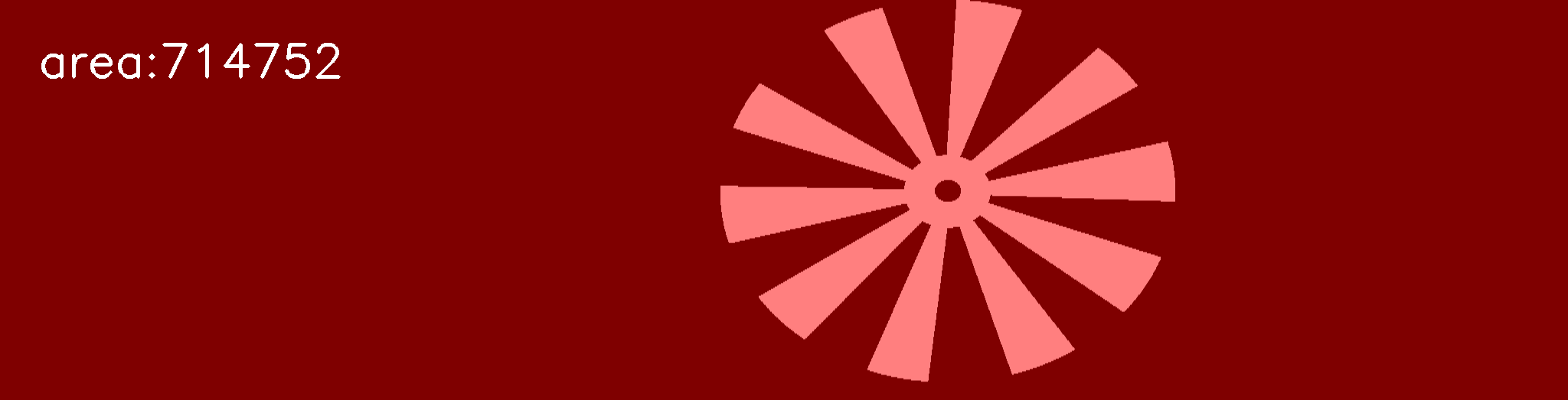}
  } \\

  \subfloat[]{
    \includegraphics[width=0.6\columnwidth]{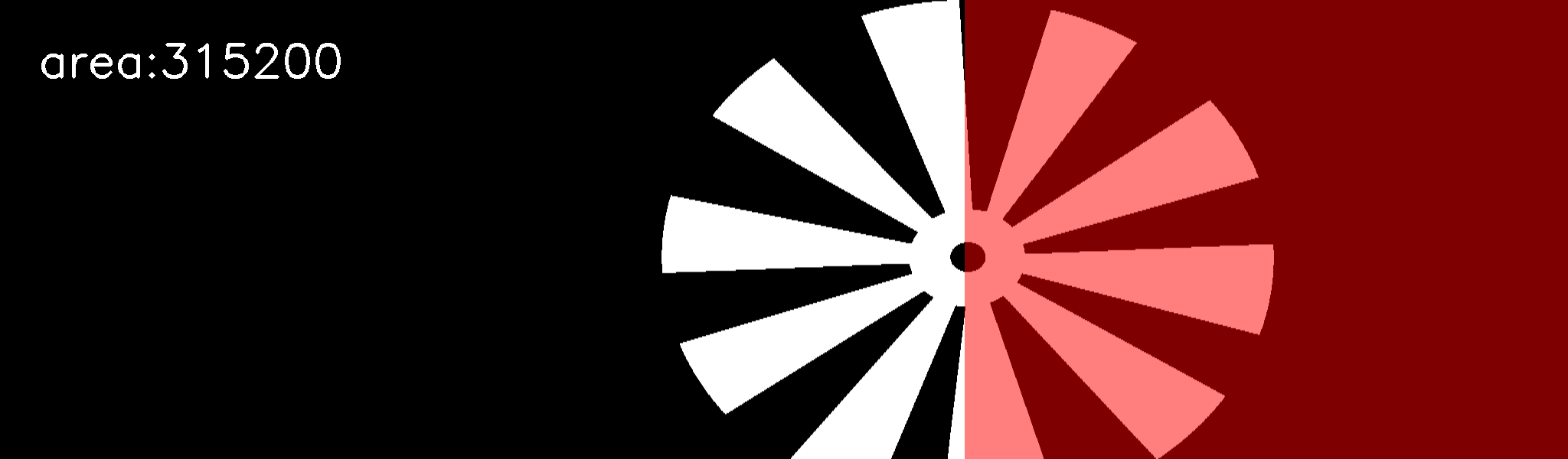}
  }
  \subfloat[]{
    \includegraphics[width=0.6\columnwidth]{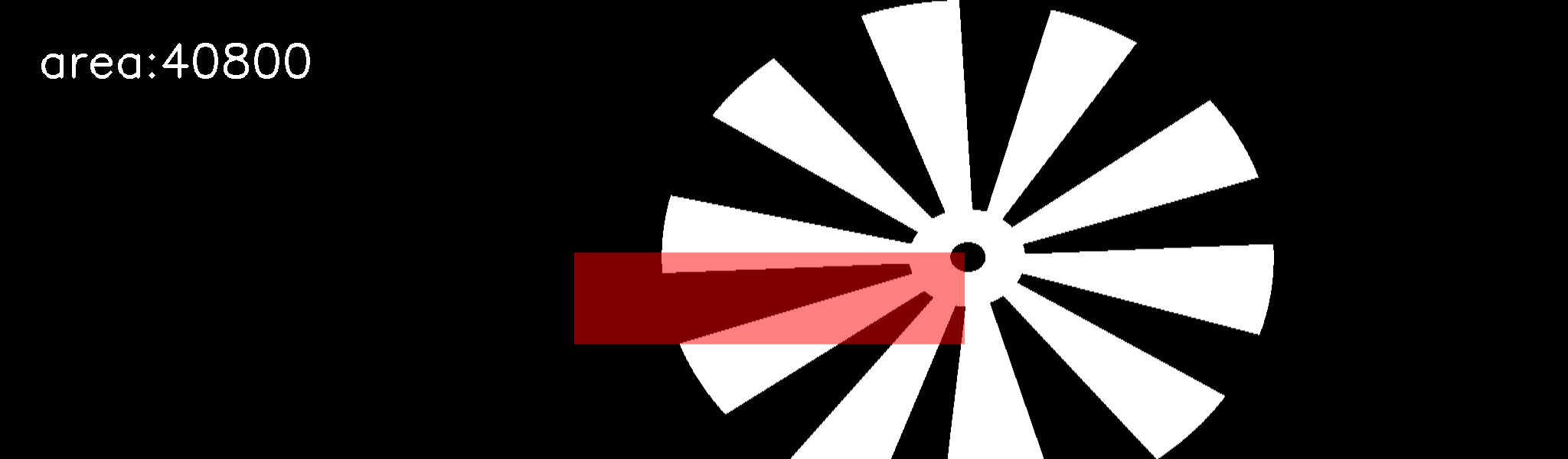}
  }
  \subfloat[]{
    \includegraphics[width=0.6\columnwidth]{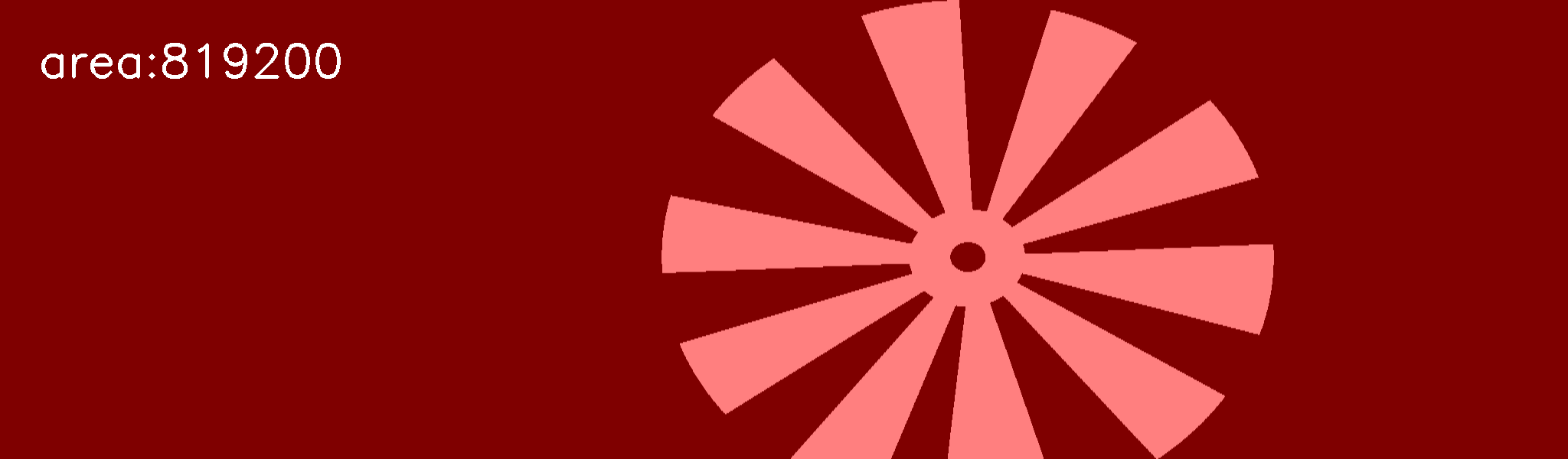}
  } \\

  \caption{Schematic diagram of dataset partitioning. Approximately 40\% of the data were used for training the network, as shown in (a)(d)(g)(j) for depths of 20 m, 15 m, 13 m, and 10 m, respectively. Around 5\% were used for $F_1$ evaluation on the validation set, as shown in (b)(e)(h)(k). The full dataset was used for visualizing imaging results, as shown in (c)(f)(i)(l).}
  \label{fig:dataset_split}
\end{figure*}

\subsection{Datasets}

\begin{figure}[!h]
  \centering
  \subfloat[]{
    \includegraphics[width=0.45\columnwidth,trim=0 50 0 100,clip]{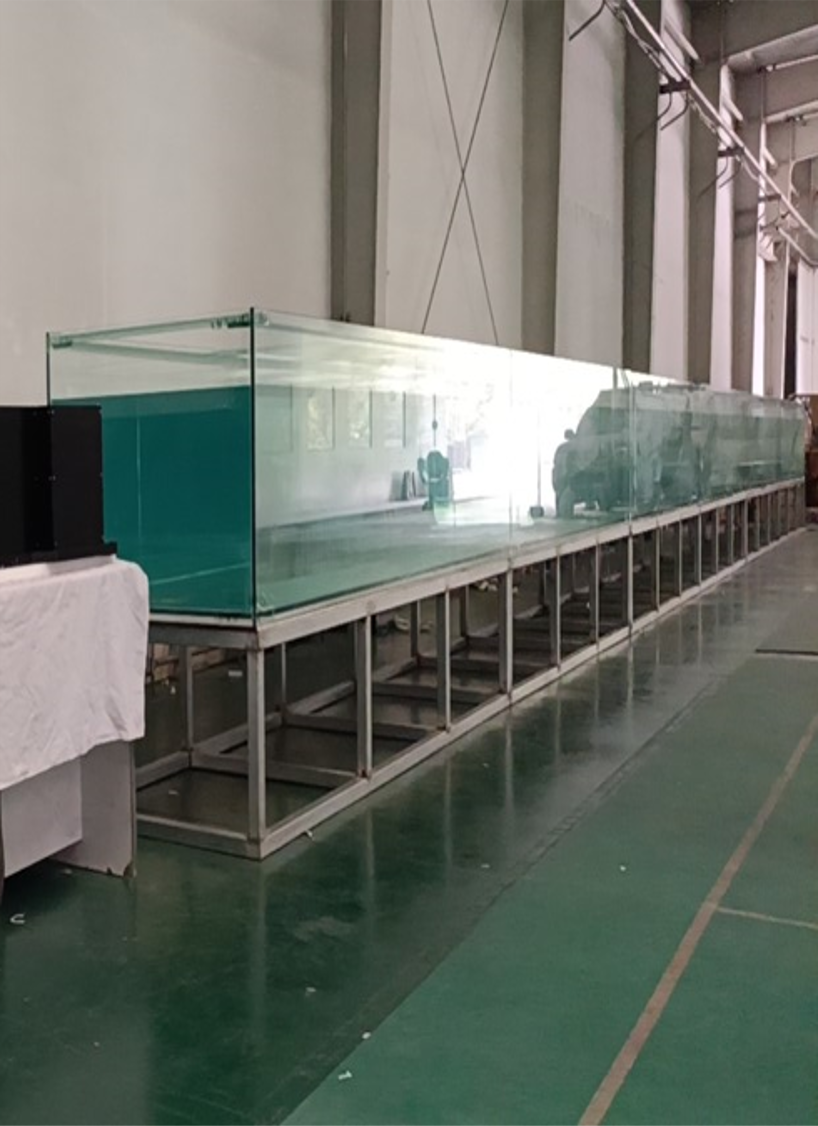}
    \label{fig:water_tank}
  }
  \subfloat[]{
    \includegraphics[width=0.45\columnwidth]{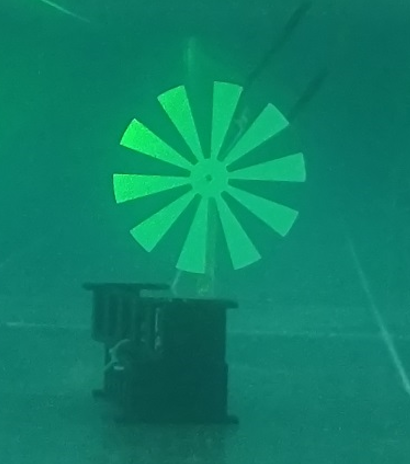}
    \label{fig:target}
  } \\
  \subfloat[]{
    \includegraphics[width=0.9\columnwidth]{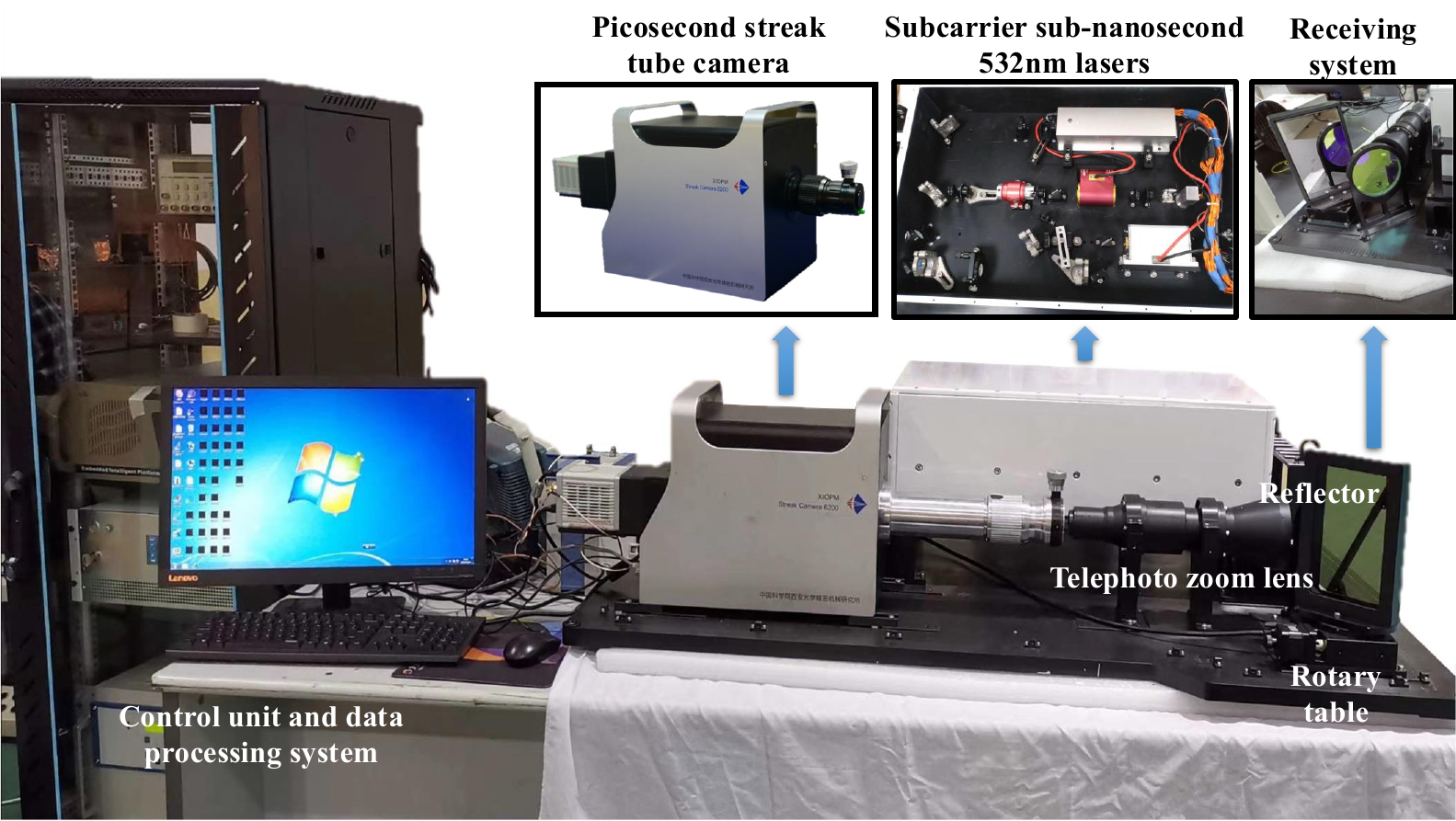}
    \label{fig:prototype}
  }
  \caption{Experimental setup. (a) The 25 m experimental water tank. (b) The 30 cm diameter experimental target. (c) Prototype of the UCLR system. }
  \label{fig:setup}
\end{figure}

The dataset was collected in a controlled environment, a 25 m long water tank (Fig. \ref{fig:water_tank}). A target with a diameter of 30 cm (Fig. \ref{fig:target}) was positioned at varying distances (10 m, 13 m, 15 m, and 20 m) within the tank, and data was collected using the self-developed UCLR system (Fig. \ref{fig:prototype}). $N_d$ discrete angles were captured at each distance, and the resolution of the streak-tube images was 2048$\times$2048. Since each row vector of the image serves as the input unit for the algorithm, each image can provide 2048 samples. $N_d$ images captured at distance $d$ generate a total of 2048$\cdot N_d$ samples, for example, we collected $N_{\text{20m}}=267$ images at the distance of 20 m, then we could have $2048\times267=546,816$ samples. These $2048\times267$ samples were manually annotated into a $2048\times267$ binary map, with each pixel assigned a value of either 0 or 1. Here, a pixel value of 0 indicates that the corresponding sample signal comprises background noise, whereas a value of 1 signifies that the sample signal contains target echoes. 

\begin{table}[htbp]
  \begin{center}
  \caption{Details of the dataset.}
  \label{tab_dataset}
  \begin{tabular}{c|ccccc}
  \toprule
  \thead{$d$} & \thead{Resolution} & \thead{$N_d$} & \thead{Test \\ set} & \thead{Training \\ set} & \thead{Validation \\ set} \\
  \midrule
  10 m & 2048$\times$2048 & 400 & 819,200 & 315,200 & 40,800 \\
  13 m & 2048$\times$2048 & 349 & 714,752 & 281,992 & 47,530 \\ 
  15 m & 2048$\times$2048 & 300 & 614,400 & 245,400 & 39,200 \\ 
  20 m & 2048$\times$2048 & 267 & 546,816 & 229,086 & 31,240 \\
  Total & 2048$\times$2048 & 1316 & 2,695,168 & 1,071,678 & 158,770 \\ 
  \bottomrule
  \end{tabular}
  \end{center}
\end{table}

Subsequently, the samples were manually divided into different subsets: approximately 40\% were allocated to the training set, which is highlighted in red in Fig. \ref{fig:dataset_20m_train}. This subset was used for network training. About 5\% of the samples were designated as the validation set, utilized for periodic evaluation of network performance during training to ensure that the best checkpoint was saved. This validation set was also utilized for performance comparison between StreakNets, StreakNets-Emb, and traditional imaging methods, as highlighted in Fig. \ref{fig:dataset_20m_valid}. Using only 5\% for validation was intended to expedite the training process. This validation subset was carefully selected to include noise and target samples that were isolated from the training dataset, ensuring representativeness. To ensure comprehensive visualization, all data samples (100\%) were designated for a final test set. This set served solely for the creation of the final image visualizations depicted in the red area of Fig. \ref{fig:dataset_20m_test} and was explicitly excluded from the performance evaluation metrics.

The partitioning method at other distances was similar to that at 20 m. In total, our dataset included 2,695,168 samples. Table \ref{tab_dataset} provides a breakdown of the number of images captured at each distance, the aggregate number of samples, and their allocation into the training and validation datasets.

\section{Experiments}

\subsection{Model training}

\begin{table}[htbp]
  \begin{center}
  \caption{Model Size and Computational Complexity of \\ Trained Models.}
  \label{tab_training}
  \begin{tabular}{c|cccc}
  \toprule
  \thead{Model Name} & \thead{Model Size} & \thead{Computational Complexity} \\
  \midrule
  StreakNet-s & 1.09 M & 2.40 GFLOPs  \\
  StreakNet-m & 2.35 M & 5.44 GFLOPs  \\
  StreakNet-l & 6.24 M & 17.19 GFLOPs  \\ 
  StreakNet-x & 25.05 M & 85.83 GFLOPs  \\
  \midrule
  StreakNetv2-s & 1.12 M & 2.40 GFLOPs  \\
  StreakNetv2-m & 2.61 M & 5.44 GFLOPs  \\ 
  StreakNetv2-l & 8.35 M & 17.19 GFLOPs  \\ 
  StreakNetv2-x & 41.87 M & 85.83 GLOPs  \\
  \midrule 
  MP-s & 1.16 M & 2.46 GFLOPs \\
  MP-m & 2.34 M & 4.90 GFLOPs \\
  \midrule 
  CNN-s & 1.06 M & 2.26 GFLOPs \\
  CNN-m & 2.08 M & 4.36 GFLOPs \\
  \bottomrule
  \end{tabular}
  \end{center}
\end{table}

Under the StreakNet-Arch, we trained StreakNet with the Self-Attention mechanism as the backbone network and StreakNetv2 with the DWC-Attention mechanism for experiments. For comparison, learning-based methods such as the MP networks \cite{Chakraverty2019,illig2020machine} and convolutional neural networks (CNN) \cite{6795724} of different scales were also trained concurrently.

During the training phase, the Stochastic Gradient Descent (SGD) algorithm is used to optimize for 120 epochs, with a base learning rate of $2\times10^{-6}$ per batch. A cosine annealing learning rate strategy is employed, and the Exponential Moving Average (EMA) method is used. The training was performed on a single NVIDIA RTX 3090 (24G). The details of all trained models are shown in Table \ref{tab_training}.

% \begin{figure}[htbp]
%   \centering
%   \includegraphics[width=0.9\columnwidth,trim=50 40 55 55,clip]{figs/lr_streaknet.pdf}
%   \caption{Learning rate variation curve during the training phase.}
%   \label{fig:lr_streaknet}
% \end{figure}

% \begin{figure}[htbp]
%   \centering
%   \includegraphics[width=0.9\columnwidth,trim=50 40 55 55,clip]{figs/loss_streaknet.pdf}
%   \caption{Loss variation curve during the training phase.}
%   \label{fig:loss_streaknet}
% \end{figure}

% \begin{figure}[htbp]
%   \centering
%   \includegraphics[width=0.9\columnwidth,trim=50 40 55 55,clip]{figs/f1_streaknet.pdf}
%   \caption{$F_1$ scores evaluated on the validation set for traditional algorithms and StreakNet.}
%   \label{fig:f1_streaknet}
% \end{figure}

\subsection{StreakNet-Arch exhibits superior anti-scattering capabilities compared to traditional imaging methods}

To address the challenge of anti-scattering, we formulate it as a binary classification task. This approach allows us to distinguish between pure noise and signal inputs containing target echoes. The $F_1$ score, a well-established metric in classification tasks, is then employed to evaluate the model's anti-scattering effectiveness.

We will evaluate the 0-1 masks $\mathbf{\hat{M}}$ obtained from the StreakNets and the traditional imaging algorithms (see Fig. \hyperref[fig:overview]{1c,1f}) using the labels provided by the dataset as ground truth $\mathbf{M}$. The $F_1$ score is calculated as Eq. \ref{eq_f1}.

\begin{equation}
  \label{eq_f1}
  \begin{split}
    P &= \frac{\sum_i \sum_j \hat{M}_{ij} \land M_{ij}}{\sum_i \sum_j \hat{M}_{ij} \land M_{ij} + \sum_i \sum_j \hat{M}_{ij} \land \neg M{ij}} , \\ 
    R &= \frac{\sum_i \sum_j \hat{M}_{ij} \land M_{ij}}{\sum_i \sum_j \hat{M}_{ij} \land M_{ij} + \sum_i \sum_j \neg \hat{M}_{ij} \land \neg M{ij}} , \\ 
    F_1 &= \frac{2 \cdot P \cdot R}{P + R} .
  \end{split}
\end{equation}

\begin{figure}[htbp]
  \centering
  \subfloat[]{
    \includegraphics[width=0.22\columnwidth]{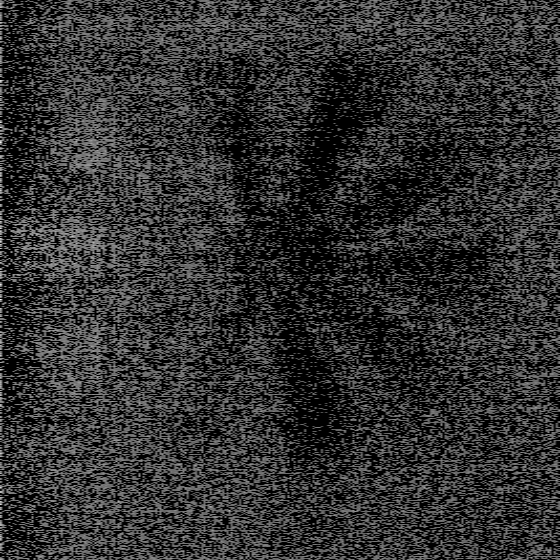}
  }
  \subfloat[]{
    \includegraphics[width=0.22\columnwidth]{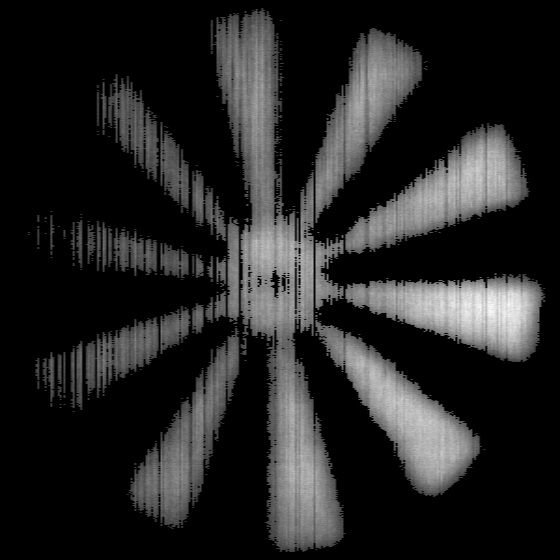}
  }
  \subfloat[]{
    \includegraphics[width=0.22\columnwidth]{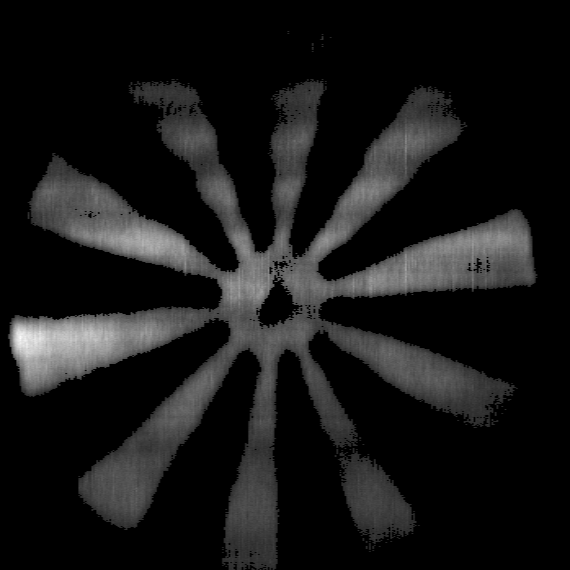}
  }
  \subfloat[]{
    \includegraphics[width=0.22\columnwidth]{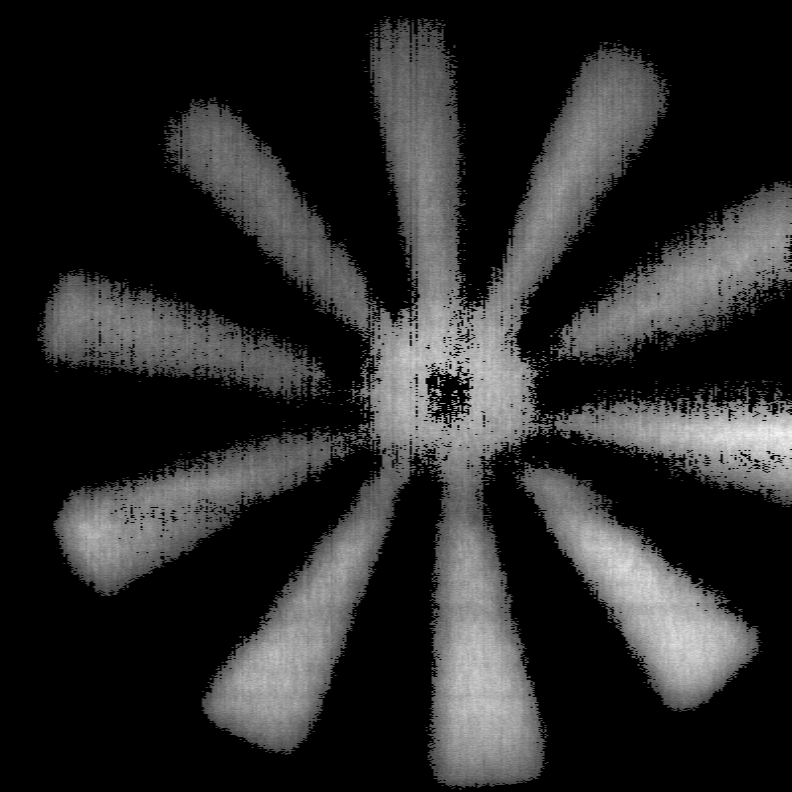}
  } \\
  
  \subfloat[]{
    \includegraphics[width=0.22\columnwidth]{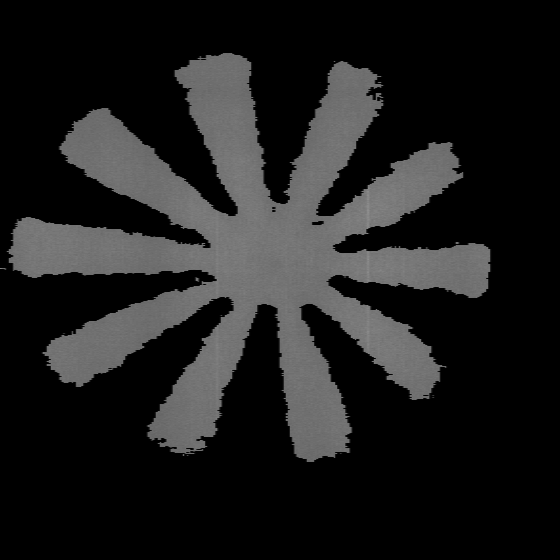}
  }
  \subfloat[]{
    \includegraphics[width=0.22\columnwidth]{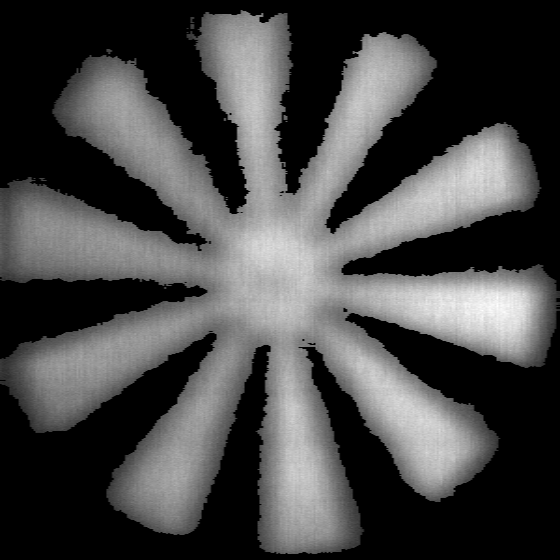}
  }
  \subfloat[]{
    \includegraphics[width=0.22\columnwidth]{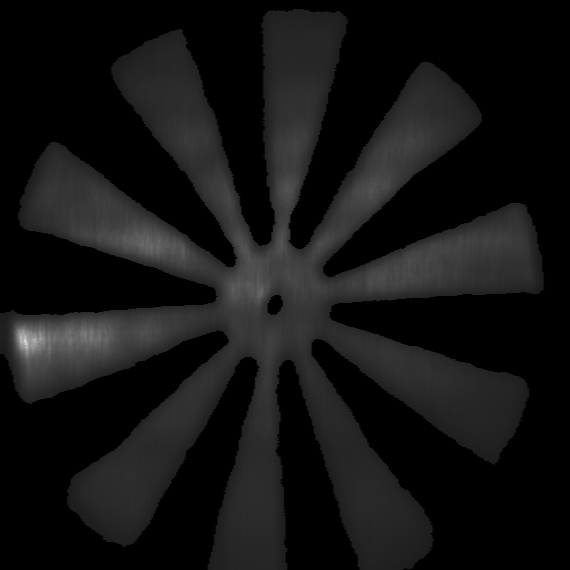}
  }
  \subfloat[]{
    \includegraphics[width=0.22\columnwidth]{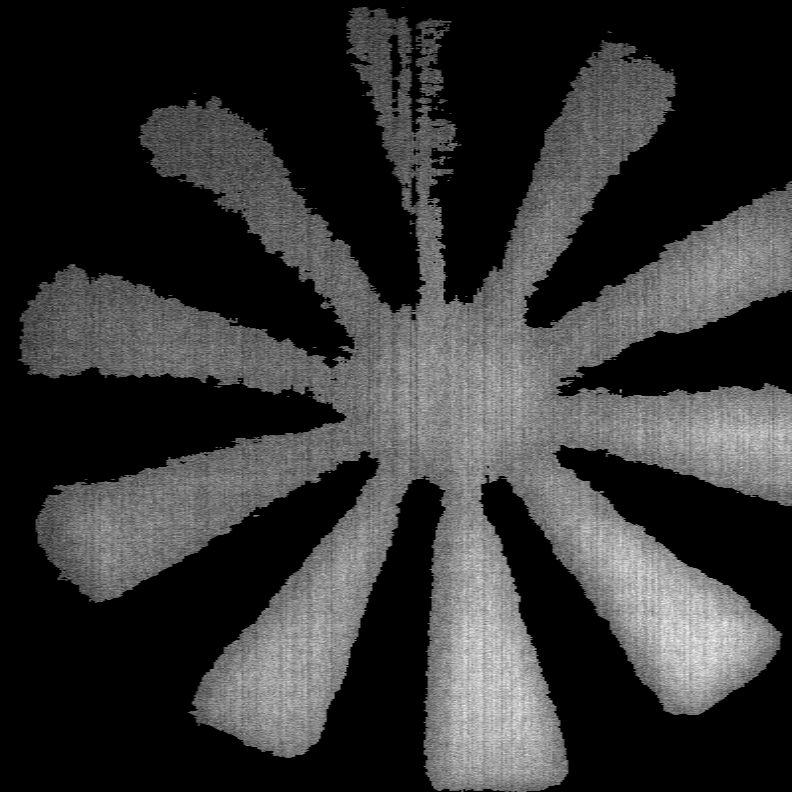}
  } \\
  
  \subfloat[]{
    \includegraphics[width=0.22\columnwidth]{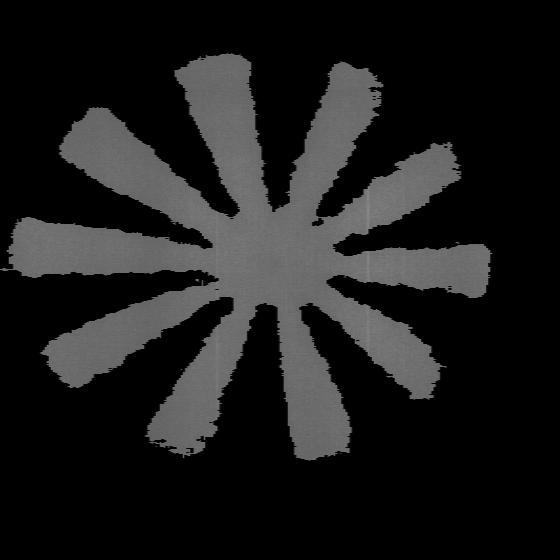}
  }
  \subfloat[]{
    \includegraphics[width=0.22\columnwidth]{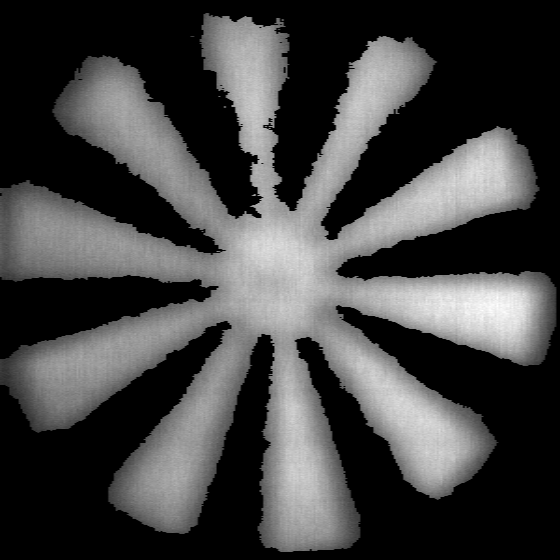}
  }
  \subfloat[]{
    \includegraphics[width=0.22\columnwidth]{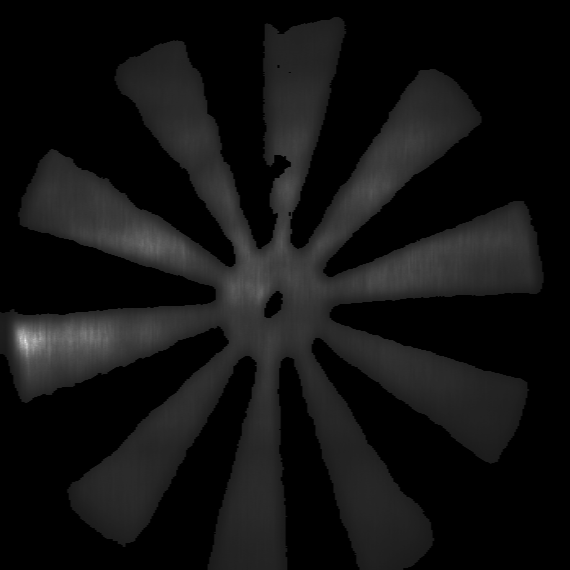}
  }
  \subfloat[]{
    \includegraphics[width=0.22\columnwidth]{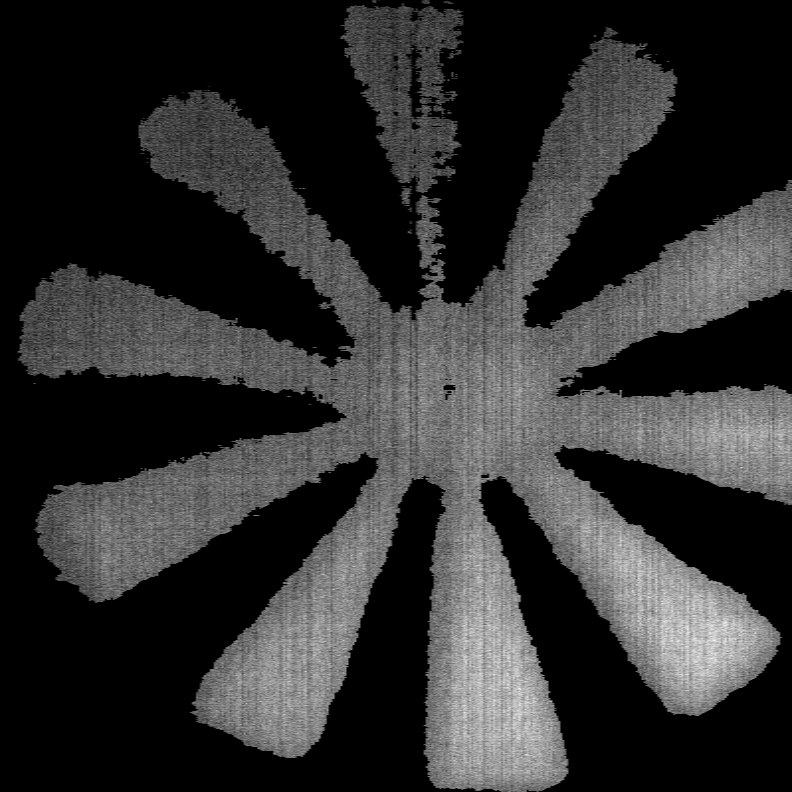}
  } \\

  \subfloat[]{
    \includegraphics[width=0.22\columnwidth]{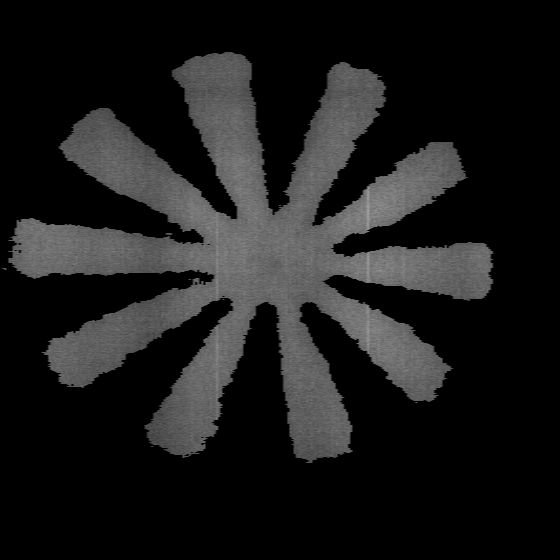}
  }
  \subfloat[]{
    \includegraphics[width=0.22\columnwidth]{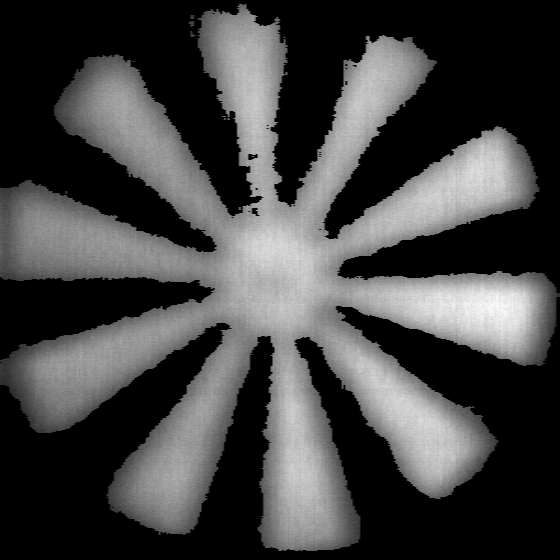}
  }
  \subfloat[]{
    \includegraphics[width=0.22\columnwidth]{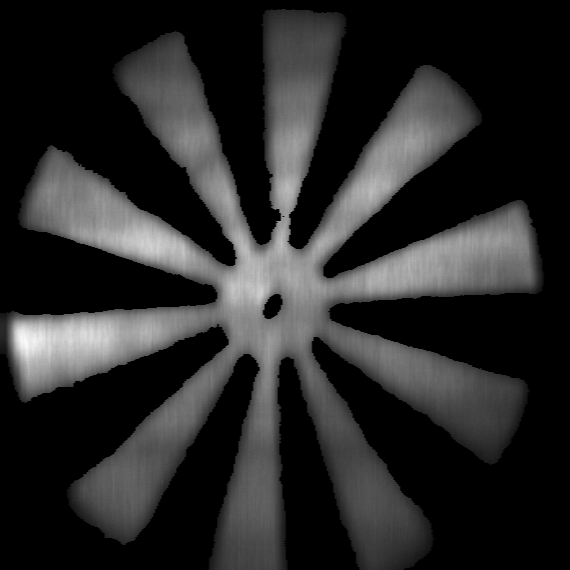}
  }
  \subfloat[]{
    \includegraphics[width=0.22\columnwidth]{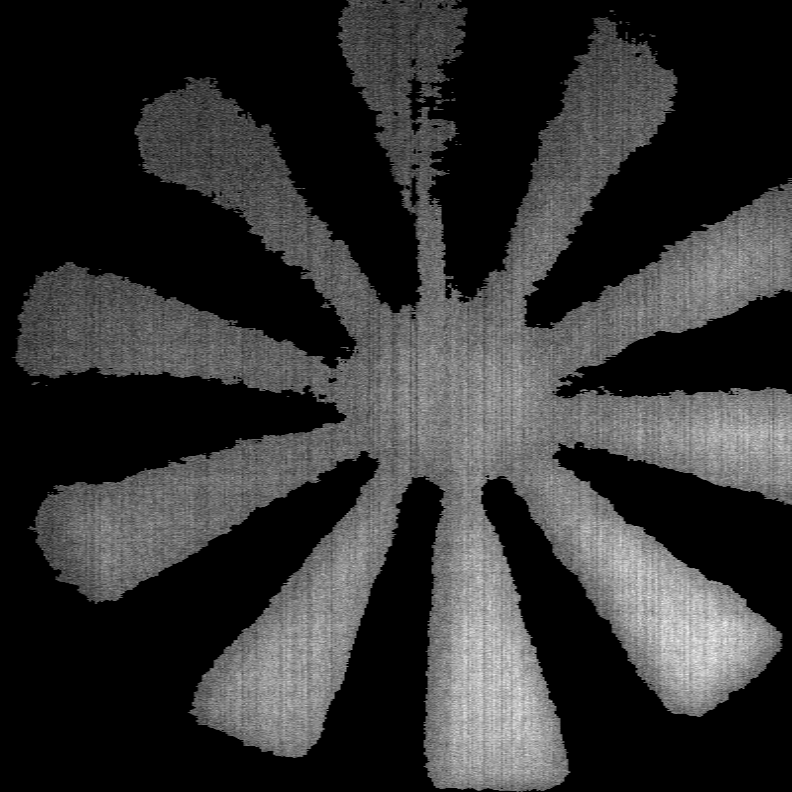}
  } \\

  \subfloat[]{
    \includegraphics[width=0.22\columnwidth]{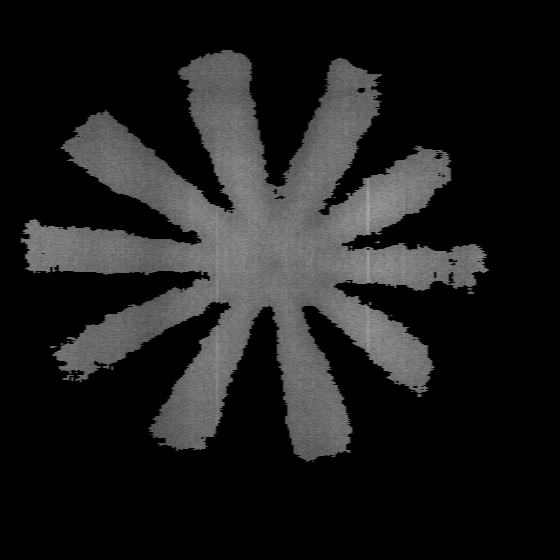}
  }
  \subfloat[]{
    \includegraphics[width=0.22\columnwidth]{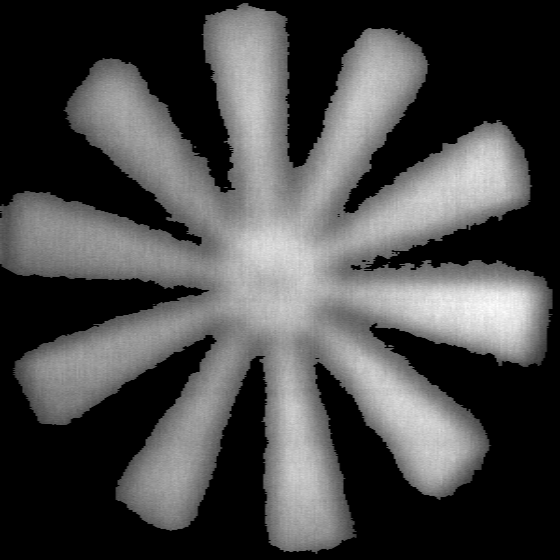}
  }
  \subfloat[]{
    \includegraphics[width=0.22\columnwidth]{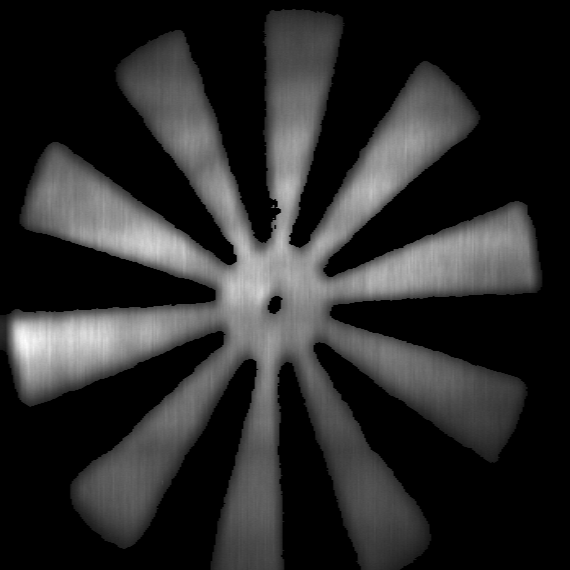}
  }
  \subfloat[]{
    \includegraphics[width=0.22\columnwidth]{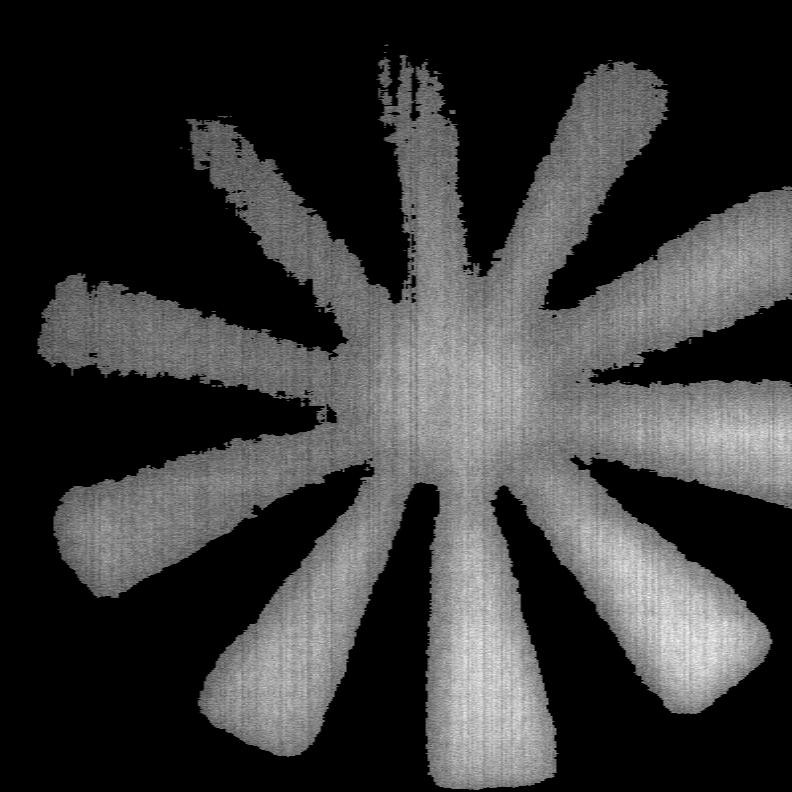}
  } \\
  
  \caption{2D imaging results at depths of 20 m, 15 m, 13 m, and 10 m for (a)–(d) Bandpass, (e)–(h) MP, (i)–(l) CNN, (m)–(p) StreakNet, and (q)–(t) StreakNetv2.}
  \label{fig:2d_results}
\end{figure}

\begin{figure}[htbp]
  \centering
  \subfloat[]{
    \includegraphics[width=0.22\columnwidth,trim=90 0 31 0,clip]{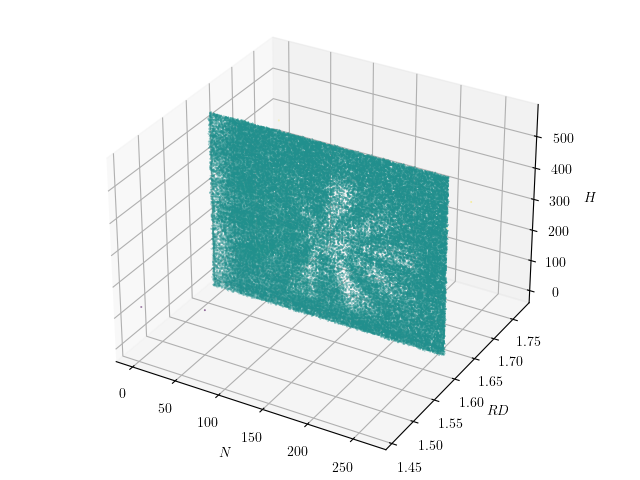}
  }
  \subfloat[]{
    \includegraphics[width=0.22\columnwidth,trim=90 0 31 0,clip]{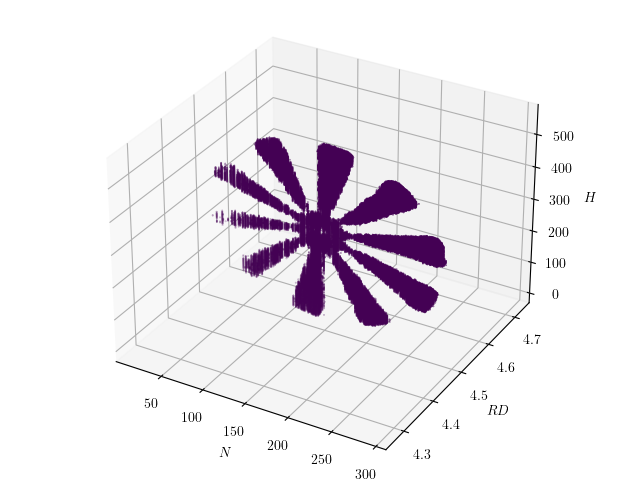}
  }
  \subfloat[]{
    \includegraphics[width=0.22\columnwidth,trim=90 0 31 0,clip]{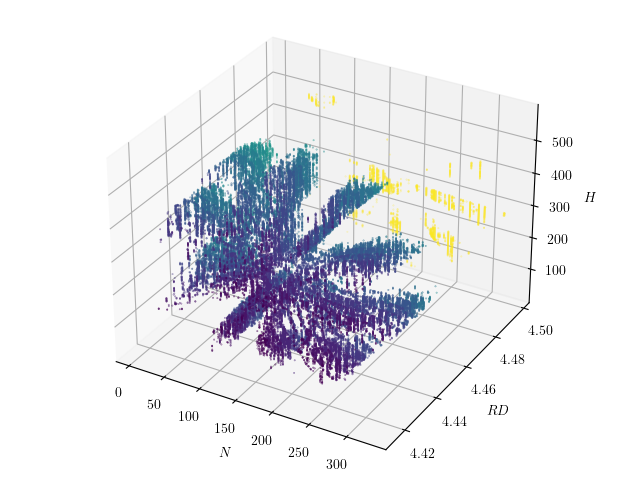}
  }
  \subfloat[]{
    \includegraphics[width=0.22\columnwidth,trim=90 0 31 0,clip]{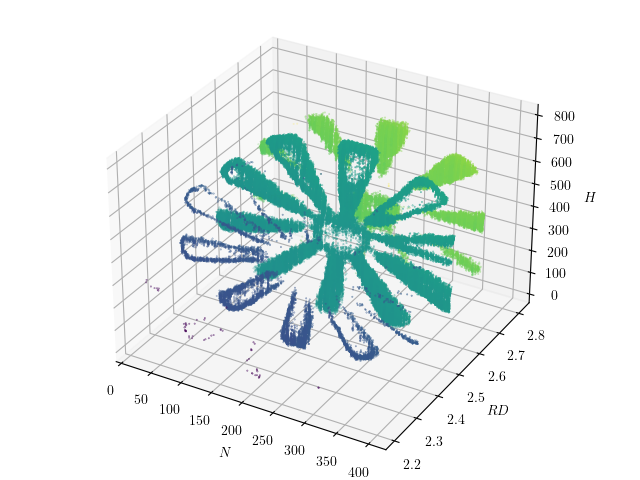}
  } \\

  \subfloat[]{
    \includegraphics[width=0.22\columnwidth,trim=90 0 31 0,clip]{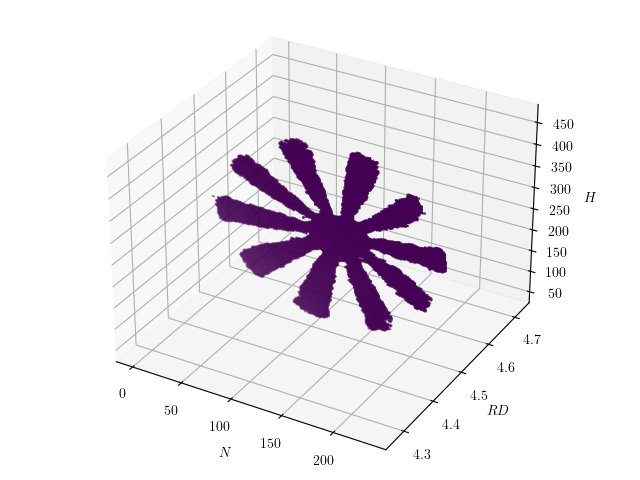}
  }
  \subfloat[]{
    \includegraphics[width=0.22\columnwidth,trim=90 0 31 0,clip]{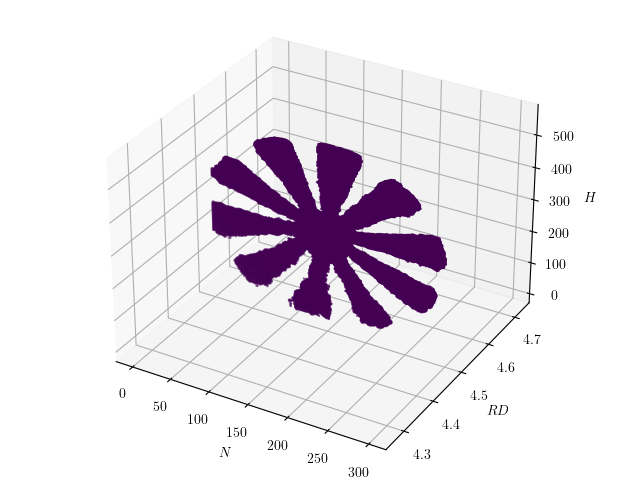}
  }
  \subfloat[]{
    \includegraphics[width=0.22\columnwidth,trim=90 0 31 0,clip]{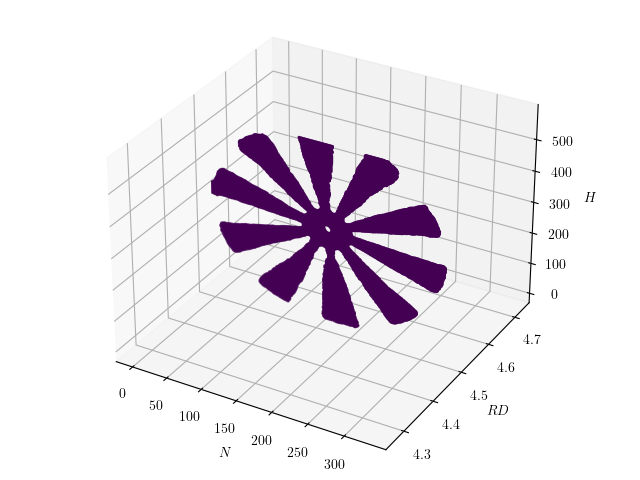}
  }
  \subfloat[]{
    \includegraphics[width=0.22\columnwidth,trim=90 0 31 0,clip]{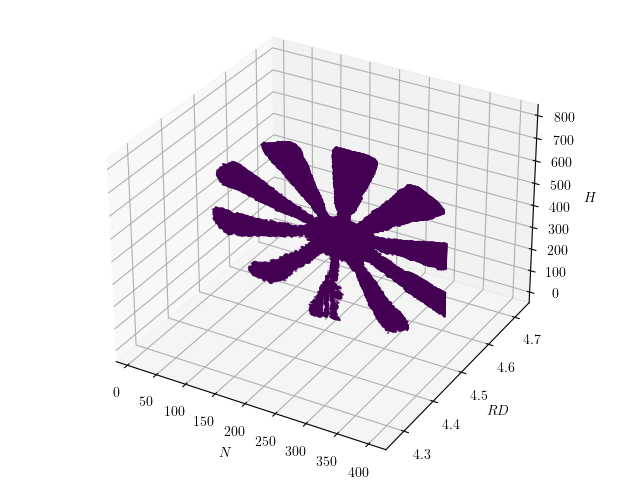}
  } \\

  \subfloat[]{
    \includegraphics[width=0.22\columnwidth,trim=90 0 31 0,clip]{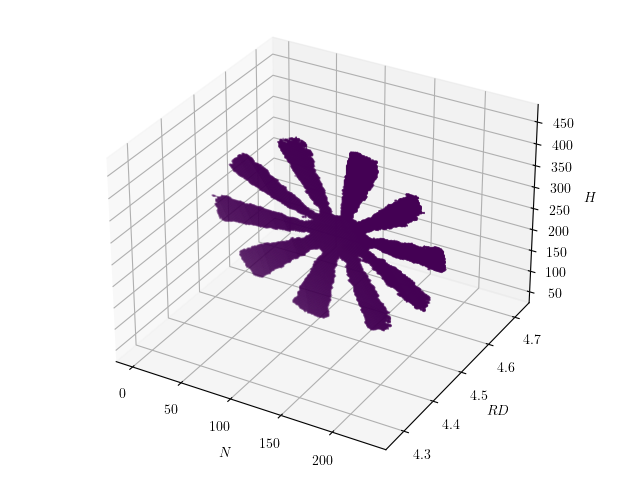}
  }
  \subfloat[]{
    \includegraphics[width=0.22\columnwidth,trim=90 0 31 0,clip]{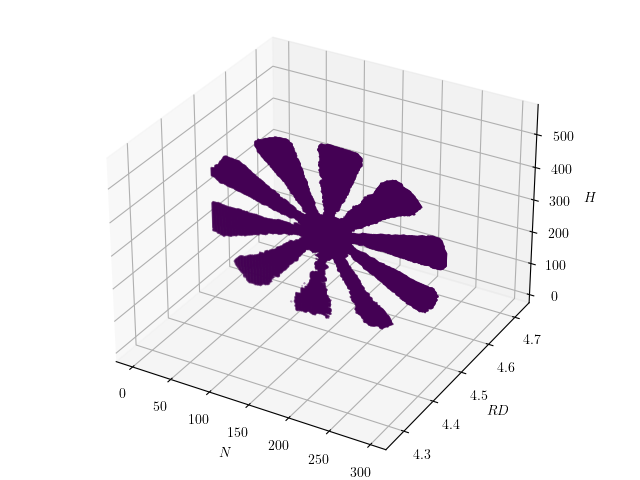}
  }
  \subfloat[]{
    \includegraphics[width=0.22\columnwidth,trim=90 0 31 0,clip]{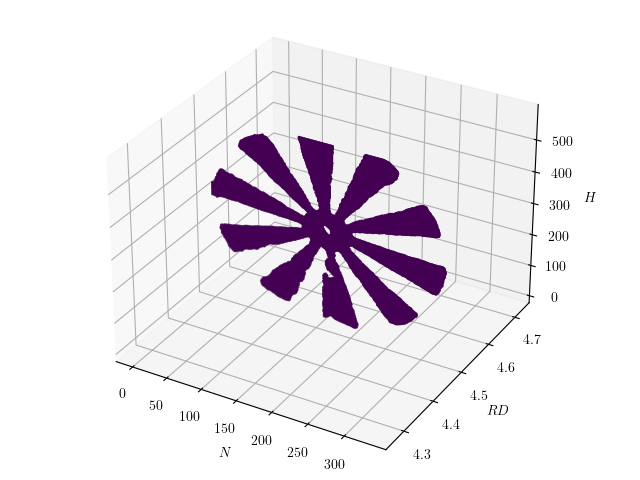}
  }
  \subfloat[]{
    \includegraphics[width=0.22\columnwidth,trim=90 0 31 0,clip]{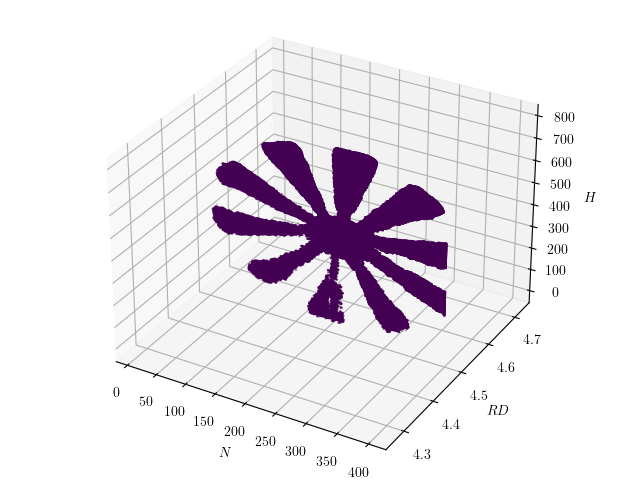}
  } \\

  \subfloat[]{
    \includegraphics[width=0.22\columnwidth,trim=90 0 31 0,clip]{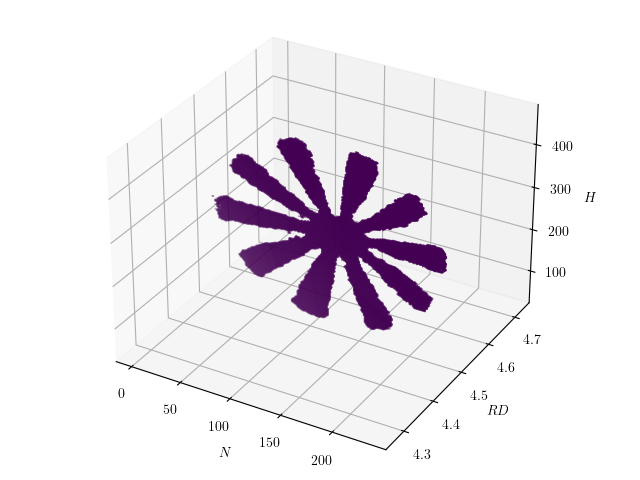}
  }
  \subfloat[]{
    \includegraphics[width=0.22\columnwidth,trim=90 0 31 0,clip]{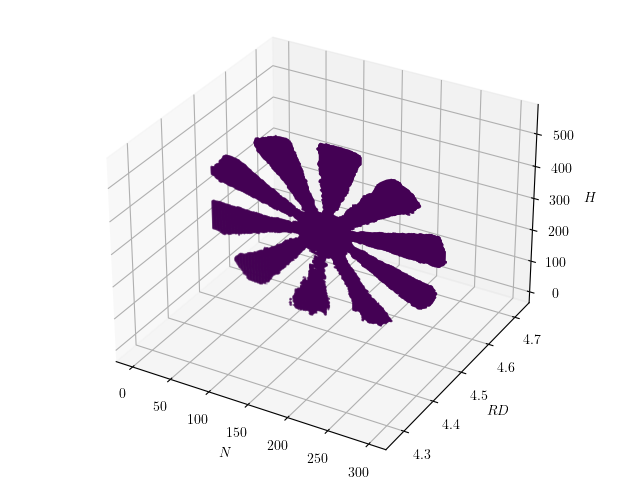}
  }
  \subfloat[]{
    \includegraphics[width=0.22\columnwidth,trim=90 0 31 0,clip]{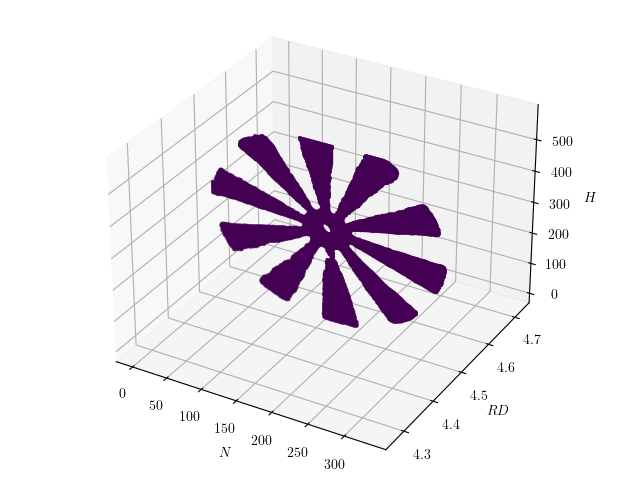}
  }
  \subfloat[]{
    \includegraphics[width=0.22\columnwidth,trim=90 0 31 0,clip]{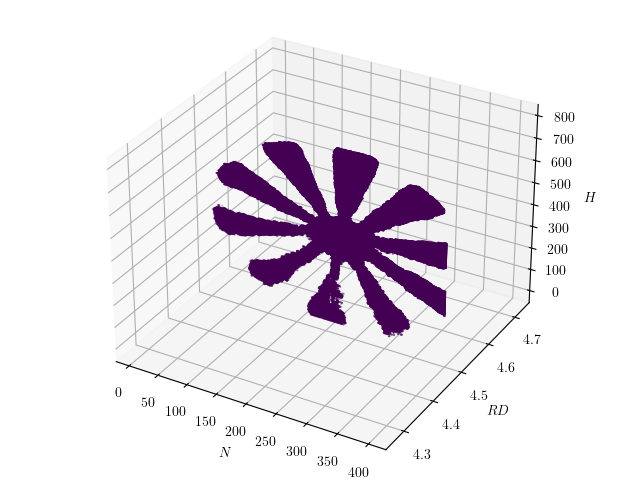}
  } \\ 
  
  \subfloat[]{
    \includegraphics[width=0.22\columnwidth,trim=90 0 31 0,clip]{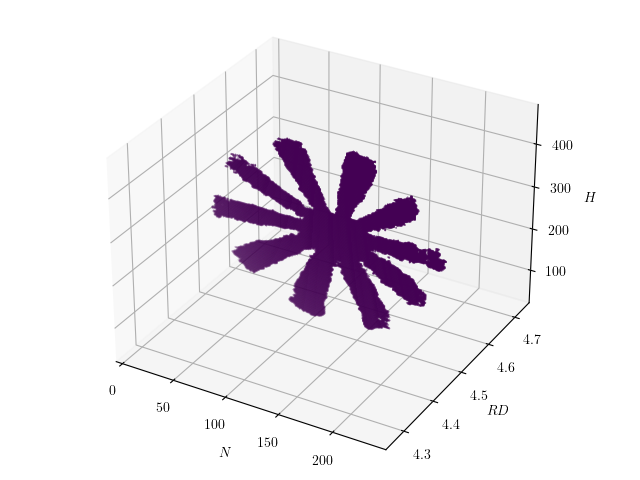}
  }
  \subfloat[]{
    \includegraphics[width=0.22\columnwidth,trim=90 0 31 0,clip]{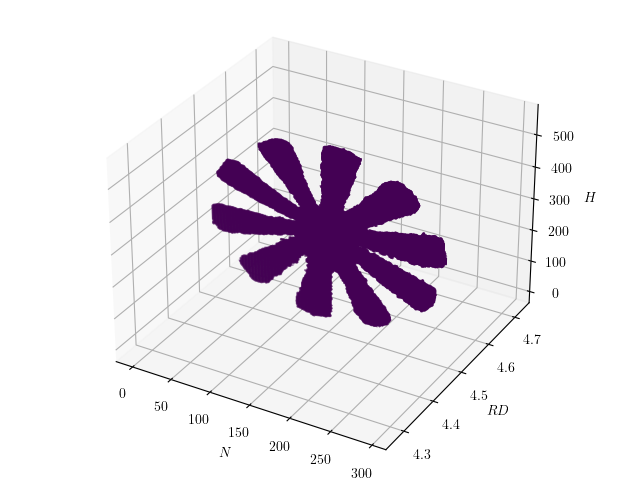}
  }
  \subfloat[]{
    \includegraphics[width=0.22\columnwidth,trim=90 0 31 0,clip]{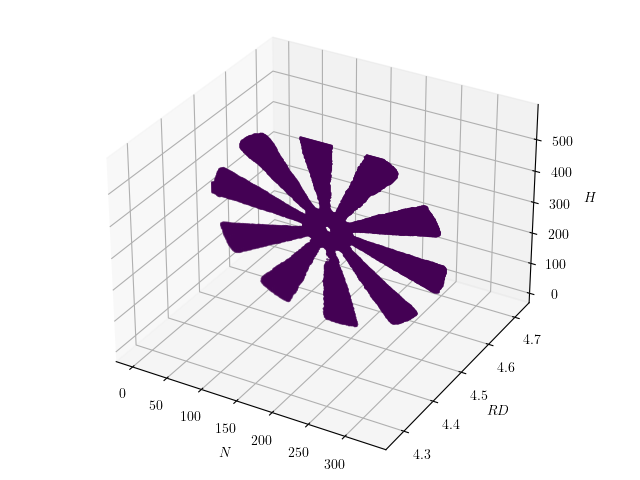}
  }
  \subfloat[]{
    \includegraphics[width=0.22\columnwidth,trim=90 0 31 0,clip]{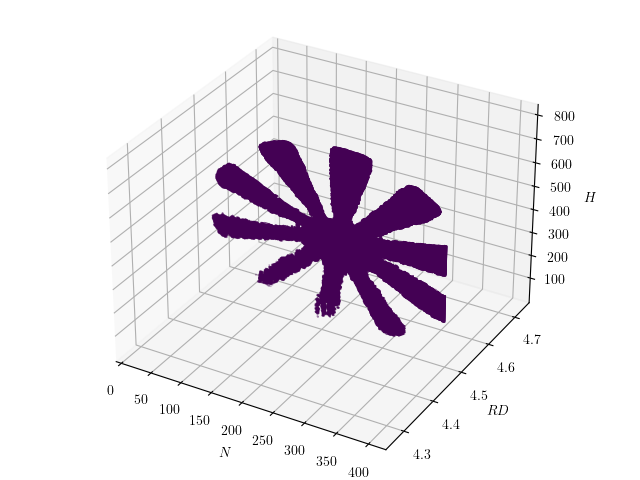}
  } \\

  \caption{3D imaging results at depths of 20 m, 15 m, 13 m, and 10 m for (a)–(d) Bandpass, (e)–(h) MP, (i)–(l) CNN, (m)–(p) StreakNet, and (q)–(t) StreakNetv2.}
  \label{fig:3d_results}
\end{figure}

Evaluation on the validation set demonstrates that both Self-Attention-based StreakNet and DWC-Attention-based StreakNetv2 significantly outperform the bandpass filtering algorithm in terms of $F_1$ score. Furthermore, with comparable model sizes and computational complexity (see Table \ref{tab_training}), models under StreakNet-Arch also achieve superior $F_1$ scores compared to learning-based MP and CNN models (Table \ref{tab_f1_streaknet}). This demonstrates that the StreakNet-Arch has stronger anti-scattering capabilities compared to traditional algorithms. The imaging results are shown in Fig. \ref{fig:2d_results} and \ref{fig:3d_results}.

\begin{table}[htbp]
  \begin{center}
  \caption{AIT (ms) for traditional imaging algorithms and StreakNet-Arch algorithms.}
  \label{tab_ait_benchmark}
  \begin{tabular}{c|cccccc}
  \toprule
  \thead{$N$} & \thead{2} & \thead{4} & \thead{8} & \thead{16} & \thead{32} & \thead{64} \\
  \midrule
  Traditional & 58.05 & 96.72 & 174.1 & 328.8 & 638.2 & 1257  \\
  \midrule
  StreakNet-s & 54.05 & 54.01 & 54.01 & 54.00 & 54.00 & 53.99  \\
  StreakNet-m & 54.89 & 54.90 & 54.92 & 54.92 & 54.93 & 54.93  \\
  StreakNet-l & 60.65 & 60.67 & 60.67 & 60.67 & 60.70 & 60.70  \\
  StreakNet-x & 84.26 & 84.28 & 84.33 & 84.30 & 84.32 & 84.33  \\
  \midrule
  StreakNetv2-s & 54.10 & 54.08 & 54.09 & 54.08 & 54.08 & 54.09  \\
  StreakNetv2-m & 55.03 & 55.05 & 55.07 & 55.08 & 55.09 & 55.09  \\
  StreakNetv2-l & 60.99 & 61.00 & 61.02 & 61.02 & 61.03 & 61.03  \\
  StreakNetv2-x & 84.11 & 84.03 & 84.03 & 84.04 & 84.08 & 84.11  \\
  \bottomrule
  \end{tabular}
  \end{center}
\end{table}

\begin{table*}[htbp]
  \begin{center}
  \caption{$F_1$ scores (\%) for traditional imaging methods and StreakNet-Arch imaging methods on the validation set.$^{\ast}$}
  \label{tab_f1_streaknet}
  \begin{tabular}{cc||cccc||cccc}
  \toprule
  \multirow{2}{*}{Baseline} & Model & StreakNet-s & StreakNet-m & StreakNet-l & StreakNet-x & StreakNetv2-s & StreakNetv2-m & StreakNetv2-l & StreakNetv2-x \\
  & $F_1(\%)$ & \textbf{86.78} & \textbf{88.23} & \textbf{86.71} & \textbf{85.57} & \textbf{86.92} & \textbf{87.03} & \textbf{86.35} & \textbf{86.33} \\
  \midrule
  Bandpass & 70.82 & \textcolor{red}{+15.96} & \textcolor{red}{+17.41} & \textcolor{red}{+15.89} & \textcolor{red}{+14.75} & \textcolor{red}{+16.10} & \textcolor{red}{+16.21} & \textcolor{red}{+15.53} & \textcolor{red}{+15.51} \\
  MP-s & 86.17 & \textcolor{red}{+0.61} & \textcolor{gray}{+2.06} & \textcolor{gray}{+0.54} & \textcolor{gray}{-0.6} & \textcolor{red}{+0.75} & \textcolor{gray}{+0.86} & \textcolor{gray}{+0.18} & \textcolor{gray}{+0.16} \\
  MP-m & 86.56 & \textcolor{gray}{+0.22} & \textcolor{red}{+1.67} & \textcolor{gray}{+0.15} & \textcolor{gray}{-0.99} & \textcolor{gray}{+0.36} & \textcolor{red}{+0.47} & \textcolor{gray}{-0.21} & \textcolor{gray}{-0.23} \\
  CNN-s & 84.89 & \textcolor{red}{+1.89} & \textcolor{gray}{+3.34} & \textcolor{gray}{+1.82} & \textcolor{gray}{+0.68} & \textcolor{red}{+2.03} & \textcolor{gray}{+2.14} & \textcolor{gray}{+1.46} & \textcolor{gray}{+1.44} \\
  CNN-m & 85.12 & \textcolor{gray}{+1.66} & \textcolor{red}{+3.11} & \textcolor{gray}{+1.59} & \textcolor{gray}{+0.45} & \textcolor{gray}{+1.80} & \textcolor{red}{+1.91} & \textcolor{gray}{+1.23} & \textcolor{gray}{+1.21} \\
  \bottomrule
  \end{tabular}
  \end{center}
  \vspace{1mm}
  {\footnotesize $^{\ast}$The $F_1$ gain of StreakNets is highlighted in red for non-learning-based methods. For learning-based methods, the red highlights indicate the $F_1$ gain of StreakNet over MP models and CNNs with comparable parameter counts and computational complexity, while all other entries are shown in gray.\par}
\end{table*}

\subsection{StreakNet-Arch is more suitable for real-time imaging than Traditional Imaging Methods}

\begin{figure}[htbp]
  \centering
  \includegraphics[width=\columnwidth]{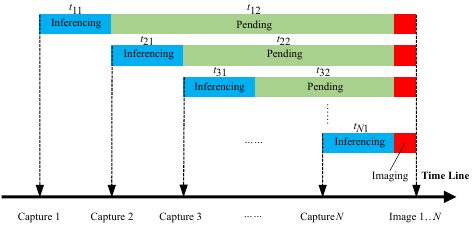}
  \caption{The sequence chart of traditional imaging algorithm. Traditional methods require all streak-tube images before thresholding and imaging can proceed, necessitating a complete wait for all data before any result is available.}
  \label{fig:time_traditional}
\end{figure}

Traditional imaging algorithms require the integration of global grayscale information to determine the denoising threshold. Therefore, for each captured streak-tube image $i(1 \leq i \leq N)$, after processing for time $t_{i1}$, there is an additional pending time $t_{i2}$ until all $N$ streak-tube images are processed. Then, additional time $t_0 \approx 0$ $(t_0 \ll t_{i1},t_{i2})$ is required to determine the threshold and complete the imaging process. Until the last streak-tube image is processed, we cannot obtain any imaging results (Fig. \ref{fig:time_traditional}).

\begin{equation}
  \label{eq_ait_traditional}
    \text{AIT}_{\text{traditional}} = \frac{N+1}{2} t_m = \frac{N+1}{2} t_m .
\end{equation}

To measure the real-time imaging capability of the algorithm, we propose an evaluation metric for the AIT, defined as the average time from the input of a streak-tube image to obtaining the corresponding imaging result. If we assume $t_{11}=t_{21}=...=t_{N1}=t_m$, the AIT for traditional imaging algorithms is calculated using Eq. \ref{eq_ait_traditional}.

\begin{figure}[htbp]
  \centering
  \includegraphics[width=\columnwidth]{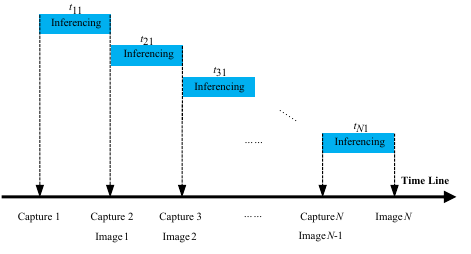}
  \caption{The sequence chart of StreakNet-Arch method. StreakNet-Arch enables immediate imaging from each input without waiting for global information.}
  \label{fig:time_streaknet}
\end{figure}

However, for the StreakNet-Arch method, there is no need for global information to determine whether the current input signal contains target echoes. Therefore, compared to traditional imaging algorithms, the StreakNet-Arch method has no pending time. Instead, for the current input streak-tube image, it can directly generate the corresponding imaging result, as shown in Fig. \ref{fig:time_streaknet}. Its AIT can be calculated using Eq. \ref{eq_ait_streaknet}.

\begin{equation}
  \label{eq_ait_streaknet}
    \text{AIT}_{\text{streaknet}} = \frac{1}{N} \sum_{i=1}^N  t_{i1} = t_m .
\end{equation}

\begin{figure}[htbp]
  \centering
  \includegraphics[width=0.9\columnwidth,trim=50 40 55 55,clip]{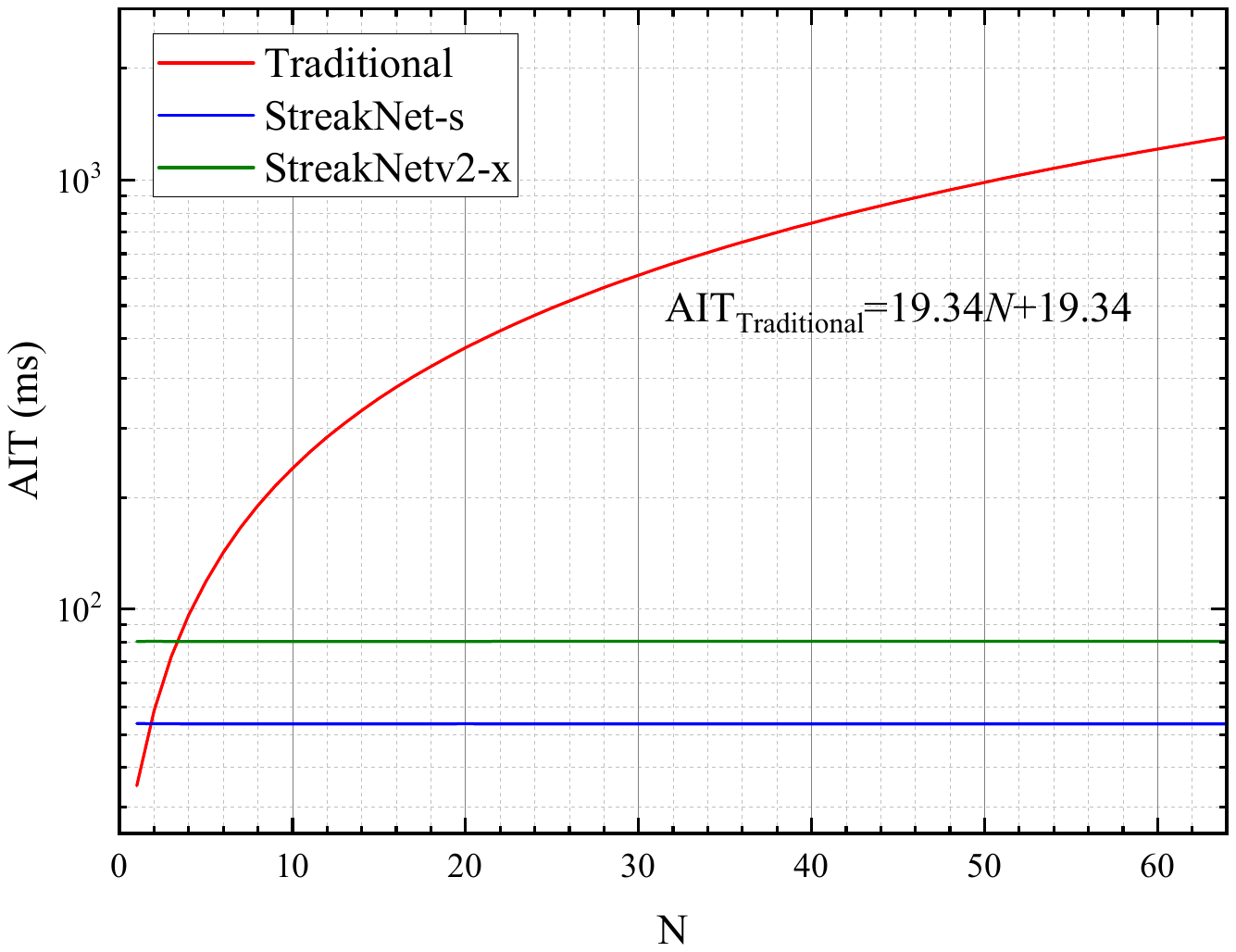}
  \caption{Curve of AIT (ms) with the changing number $N$ of streak-tube images. It is evident that the AIT of the traditional imaging algorithm is a linear function with respect to $N$, while the AIT of StreakNet-Arch is a constant.}
  \label{fig:ait_benchmark}
\end{figure}

It is evident that the AIT of the traditional imaging algorithm is a linear function with respect to $N$, while the AIT of StreakNet-Arch is a constant. Therefore, theoretically, in practical scenarios where $N$ is large, StreakNet-Arch will have a significant advantage in real-time imaging. To validate this theory, we conducted a comparative experiment. We sequentially input $N$ steak-tube images, where $N$ gradually increases from 1 to 64, and tested the AIT metric for both traditional algorithms and StreakNet-Arch on an NVIDIA RTX 3060 (12G) GPU. Experimental results are depicted in Fig. \ref{fig:ait_benchmark} and Table \ref{tab_ait_benchmark}, with AIT values measured in milliseconds (ms). The experimental findings indicate that as the number of streak-tube images increases from 2 to 64, the AIT for traditional imaging methods escalates linearly from 58 ms to 1257 ms. In contrast, the AIT for the StreakNet method remains constant within the range of 54 ms to 84 ms.

The experimental results validate the correctness of the theory: the AIT of the traditional imaging algorithm varies linearly with the number of images (the vertical axis is in logarithmic form in Fig. \ref{fig:ait_benchmark}), while the AIT of StreakNet-Arch is a constant. When $N > 4$, the AIT of StreakNet-Arch is significantly better than that of the traditional algorithm, confirming that StreakNet-Arch is more suitable for real-time imaging tasks.

\subsection{FD Embedding Layer is an equivalent bandpass filter}

\begin{figure*}[!ht]
  \centering
  \subfloat[]{
    \includegraphics[width=0.82\columnwidth,trim=50 40 55 55,clip]{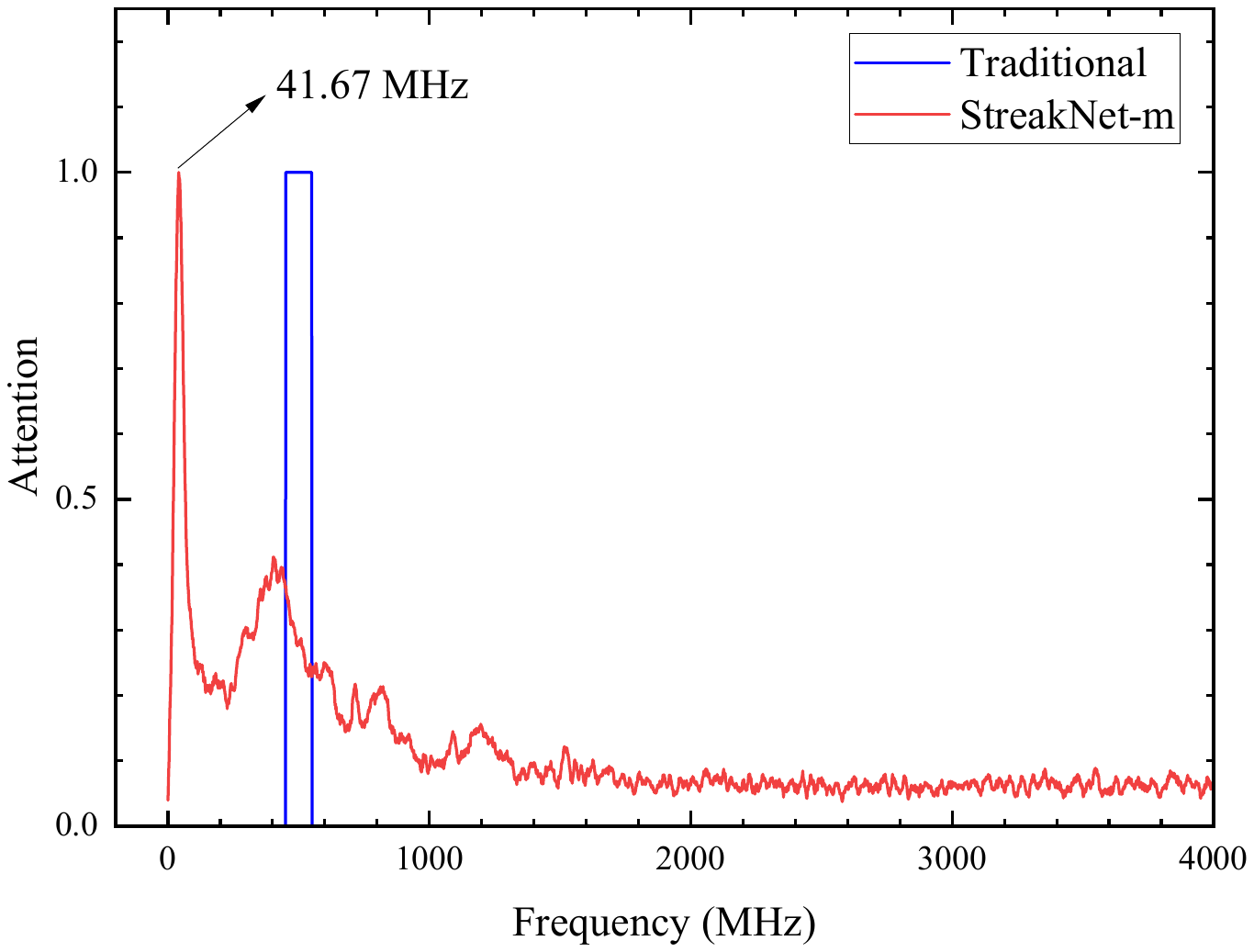}
    \label{fig:streaknet_m_embedding}
  } \hspace{0.3cm}
  \subfloat[]{
    \includegraphics[width=0.82\columnwidth,trim=50 40 55 55,clip]{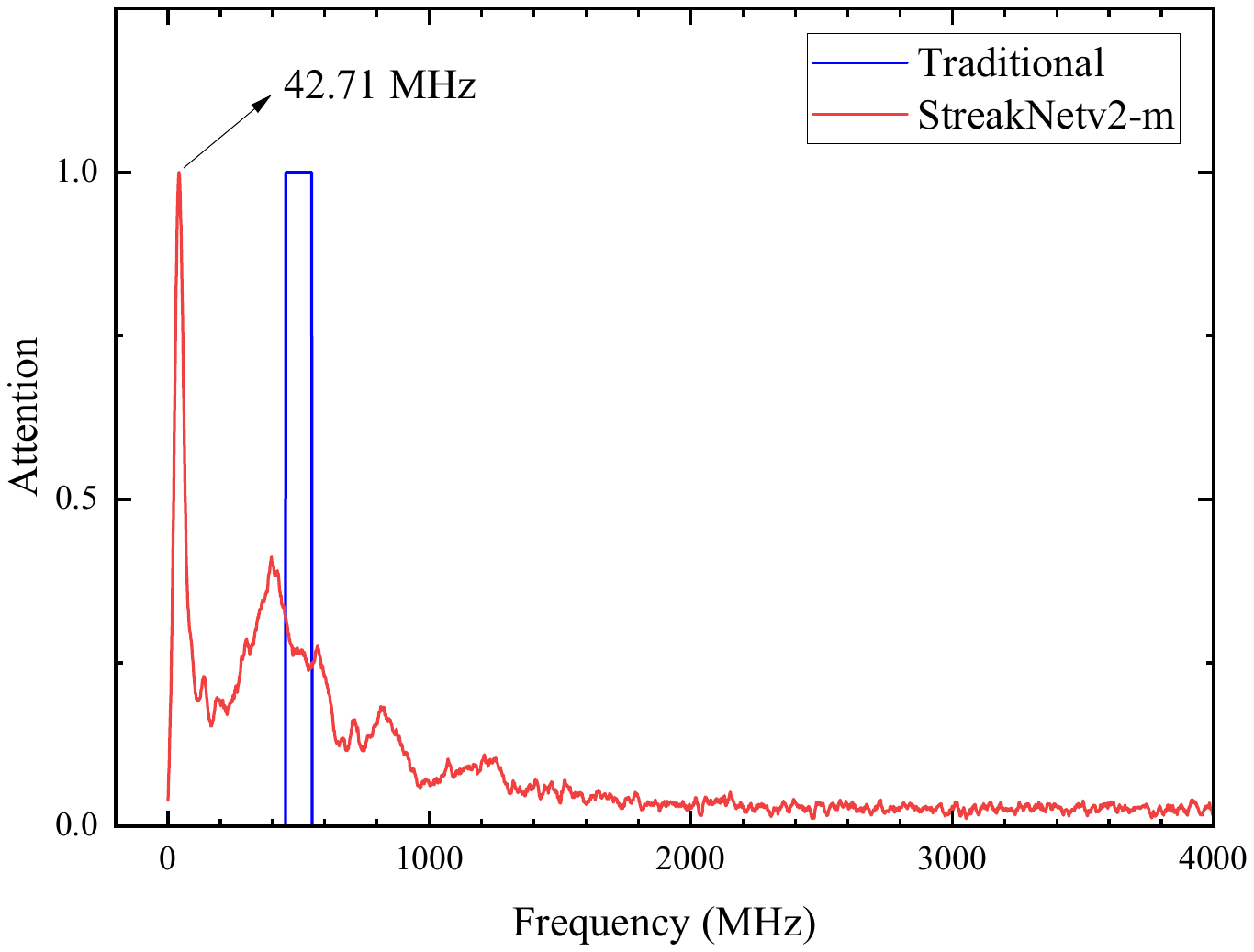}
    \label{fig:streaknetv2_m_embedding}
  }
  \caption{Results of attention distribution with frequency after AAM analysis. (a) Attention distribution with frequency after AAM analysis for StreakNet-m. (b) Attention distribution with frequency after AAM analysis for StreakNetv2-m.}
  \label{fig:streaknet_embedding}
\end{figure*}

To further explore the potential learning mechanisms of StreakNets, we performed AAM on the FD Embedding Layer of StreakNet-m and StreakNetv2-m, which performed best on the validation set, and visualized the attention distribution, as shown in Fig. \ref{fig:streaknet_embedding}.

Since the carrier frequency of the detection signal is 500 MHz (see Fig. \hyperref[fig:overview]{1e}), traditional bandpass imaging algorithms use a handcraft bandpass filter with a range of 450 MHz - 550 MHz during filtering. If we consider the bandpass filter from the perspective of ``attention distribution'', we can think of the bandpass filter as a binary attention distribution with values of 1 for frequencies in the range of 450 MHz - 550 MHz and 0 for frequencies outside this range. The FD Embedding Layer's attention distribution offers a similar concept, functioning as a learnable generalized bandpass filter.

In Fig. \ref{fig:streaknet_m_embedding} and Fig. \ref{fig:streaknetv2_m_embedding}, we observed that the FD Embedding Layer has a significant attention towards frequencies near 500 MHz, which closely matches the range of the traditional bandpass filtering algorithm within an acceptable margin of error. However, apart from frequencies near 500 MHz, the highest attention appears around 40 MHz, which seems counterintuitive.

\begin{figure*}[!ht]
  \centering
  \subfloat[]{
    \includegraphics[width=0.82\columnwidth,trim=50 40 55 55,clip]{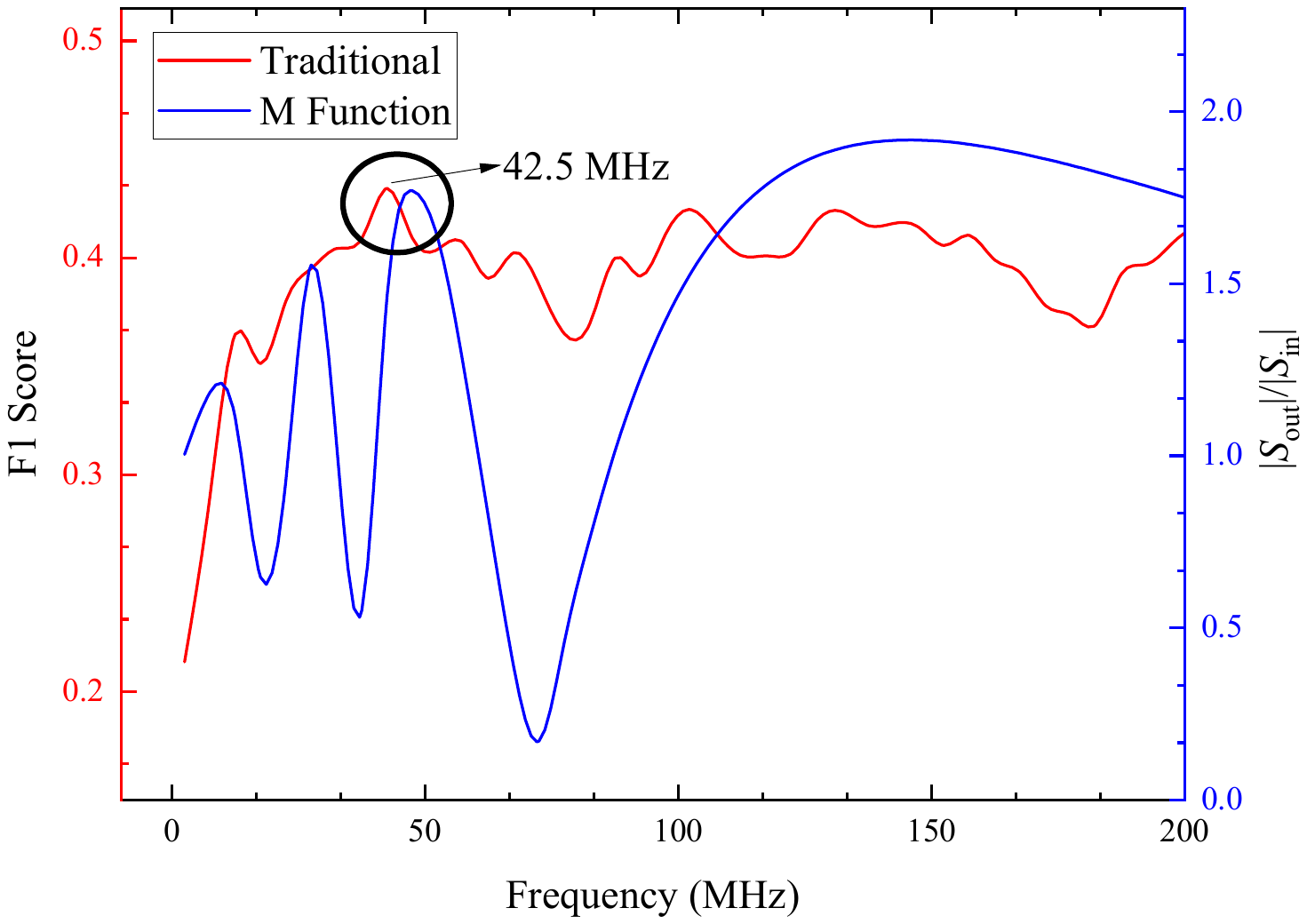}
    \label{fig:mfunc_f1}
  } \hspace{0.3cm}
  \subfloat[]{
    \includegraphics[width=0.82\columnwidth,trim=50 40 55 55,clip]{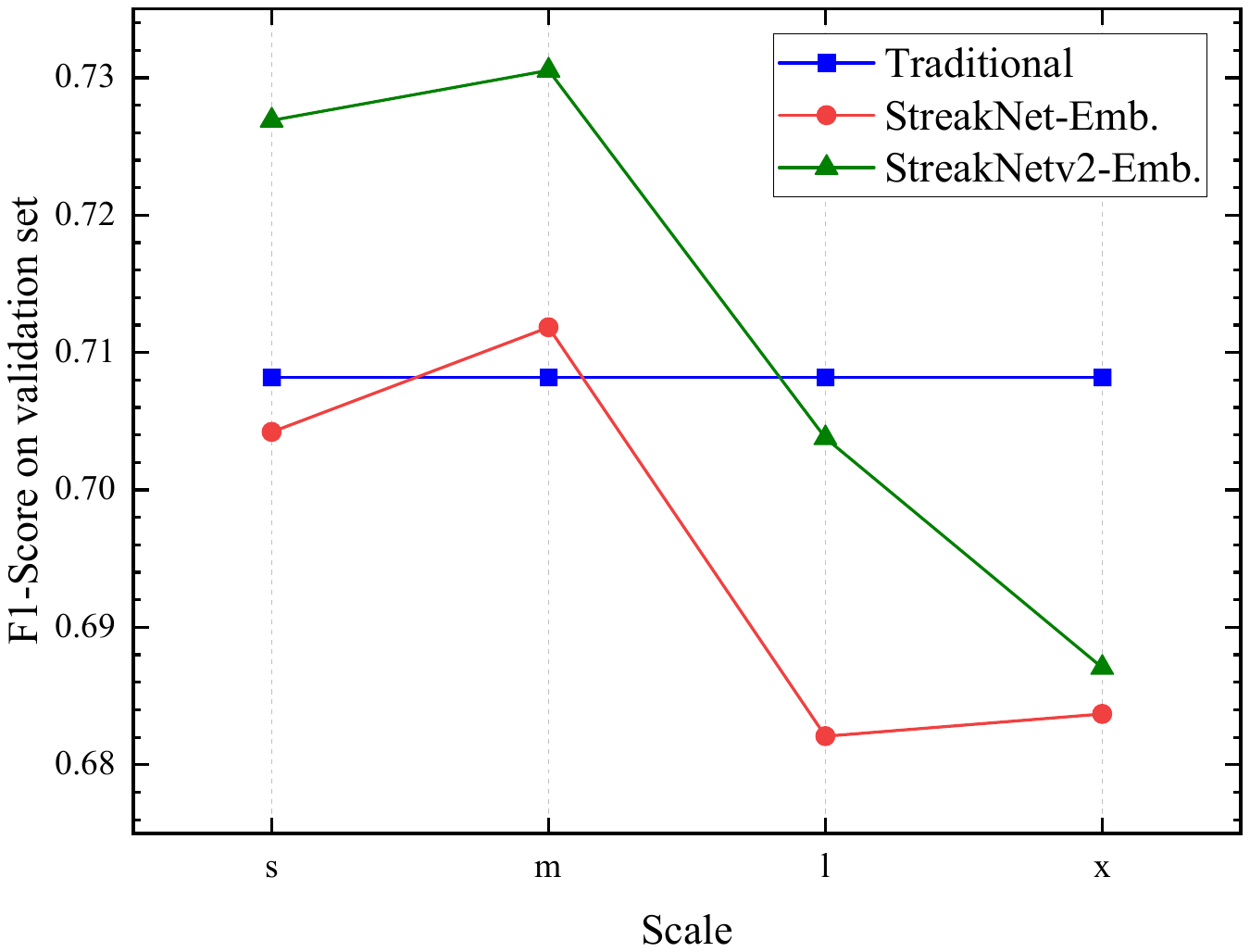}
    \label{fig:f1_streaknet_embedding}
  }
  \caption{Results of bandpass range enumeration experiment and network ablation experiment. (a) The $F_1$ scores on the validation set for imaging results using traditional bandpass filters in different frequency ranges (red curve, left), and the curve of $\mathcal{M}$ Function (blue curve, right). (b) The $F_1$ scores on the validation set for imaging using traditional bandpass filtering methods and imaging using the generalized bandpass filter obtained directly from AAM.}
  % \label{fig:streaknet_embedding}
\end{figure*}

Therefore, we further enumerated the range of bandpass filters in the range of 0 - 200 MHz, with each group spanning 5 MHz, and used traditional bandpass methods for imaging. We then calculated the $F_1$ score on the validation set. The experimental results are shown as the red curve in Fig. \ref{fig:mfunc_f1}. A peak appears at 42.5 MHz (i.e., the 40 MHz - 45 MHz bandpass range), indicating that frequency information near 40 MHz is indeed strongly correlated with anti-scattering imaging.

After consulting the literature on physical optics, we found that Perez et al. proposed a physical model for the frequency response of water in 2012 \cite{perez_techniques_2012}, called $\mathcal{M}$ Function, as shown in Eq. \ref{eq_mfunc}.

\begin{equation}
  \label{eq_mfunc}
  \mathcal{M}(\Delta Z)=\sqrt{1+e^{-2 \varepsilon \Delta Z}-2e^{- \varepsilon \Delta Z} \cos(K \Delta Z)} ,
\end{equation}

\noindent where $\mathcal{M}$ represents  the ratio of the amplitude of the output signal frequency component to that of the input signal, \textit{i.e.}, the transfer function. $\varepsilon$ is the attenuation coefficient, $\Delta Z$ is half the wavelength corresponding to the carrier frequency, and $K$ is the number of carrier pulses. 

\begin{figure*}[!ht]
  \centering
  \subfloat[]{
    \includegraphics[width=0.82\columnwidth,trim=50 40 55 55,clip]{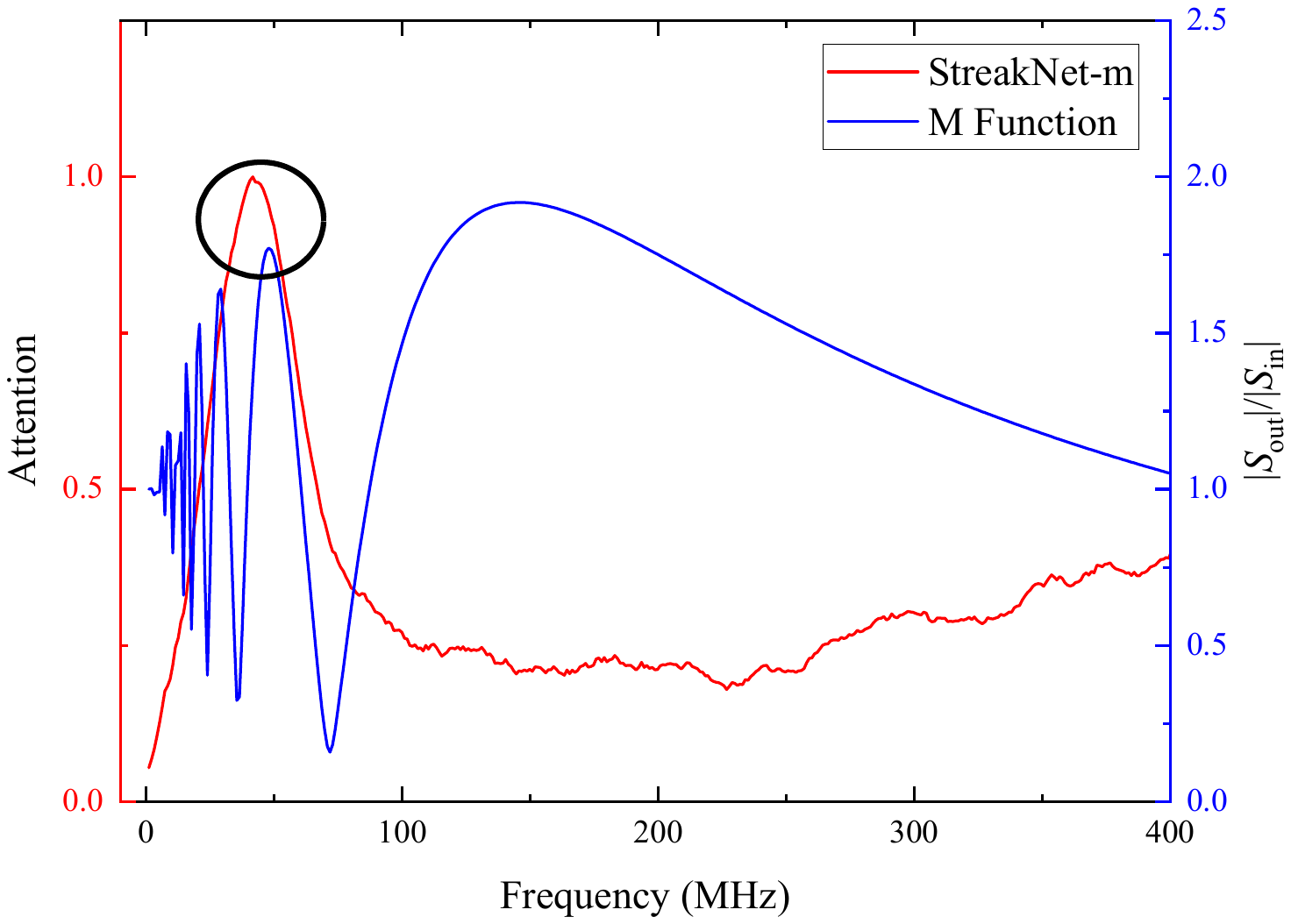}
    % \label{fig:streaknet_m_embedding}
  } \hspace{0.3cm}
  \subfloat[]{
    \includegraphics[width=0.82\columnwidth,trim=50 40 55 55,clip]{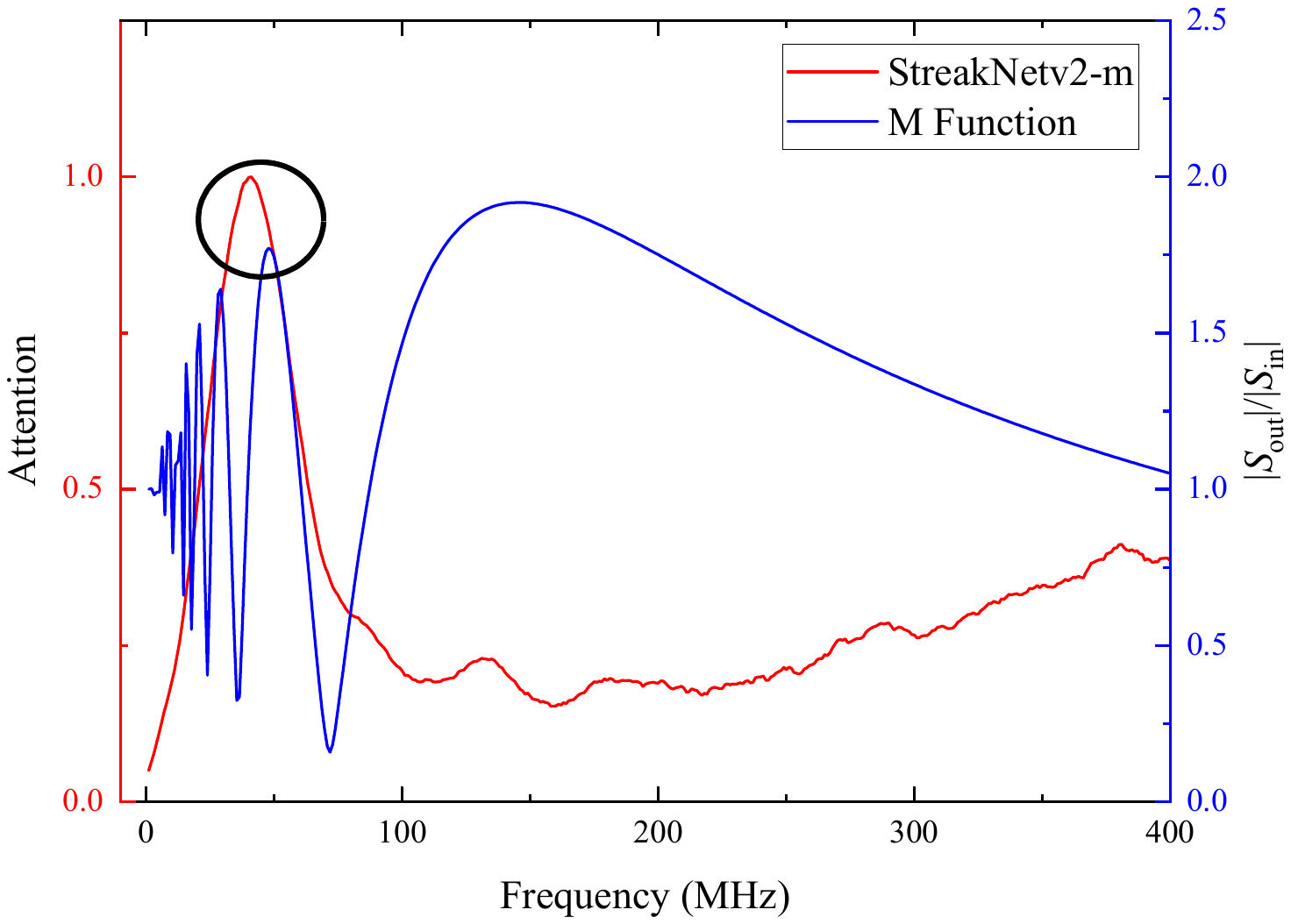}
    % \label{fig:streaknetv2_m_embedding}
  }
  \caption{The curve of Attention distribution of StreakNets and the curve of $\mathcal{M}$ Function. (a) Attention distribution of StreakNet-m and the curve of $\mathcal{M}$ Function. (b) Attention distribution of StreakNetv2-m and the curve of $\mathcal{M}$ Function.}
  \label{fig:mfunc_streaknet}
\end{figure*}

In our experiment, the attenuation coefficient $\varepsilon$ of water is 0.11, and the number of carrier pulses $K$ is 4. The $\mathcal{M}$ Function curve plotted is shown as the blue curve in Fig. \ref{fig:mfunc_f1}. And surprisingly, it is found that within an acceptable error range, there is indeed a peak near 40 MHz. By plotting the $\mathcal{M}$ Function and the attention distribution of the FD Embedding Layer on the same Fig. \ref{fig:mfunc_streaknet}, it is also found that this peak almost perfectly overlaps.

The experiments above indicate that StreakNets have learned from a large amount of sample data and discovered that frequency components near 40 MHz have a greater impact on anti-scattering imaging than those near 500 MHz. Therefore, they allocate more attention to these frequency components. Besides, the distribution obtained through AAM is a more powerful generalized bandpass filter. Although the learning mechanisms of current deep learning technologies still lack interpretability, the counterintuitive results obtained by the network may provide research insights for physical optics researchers to establish more comprehensive physical models of water bodies or guide algorithm researchers to design more advanced manual filters.

\subsection{DBC-Attention is more suitable for underwater optical 3D imaging than Self-Attention}

To demonstrate the superiority of DBC-Attention over Self-Attention in underwater imaging tasks, we conducted ablation experiments by replacing the Self-Attention module in StreakNet with DBC-Attention while keeping all other parameters unchanged. Through the experimental results Table \ref{tab_f1_streaknet}), we found:

\begin{figure*}[!h]
  \centering
  \subfloat[]{
    \includegraphics[width=0.729\columnwidth]{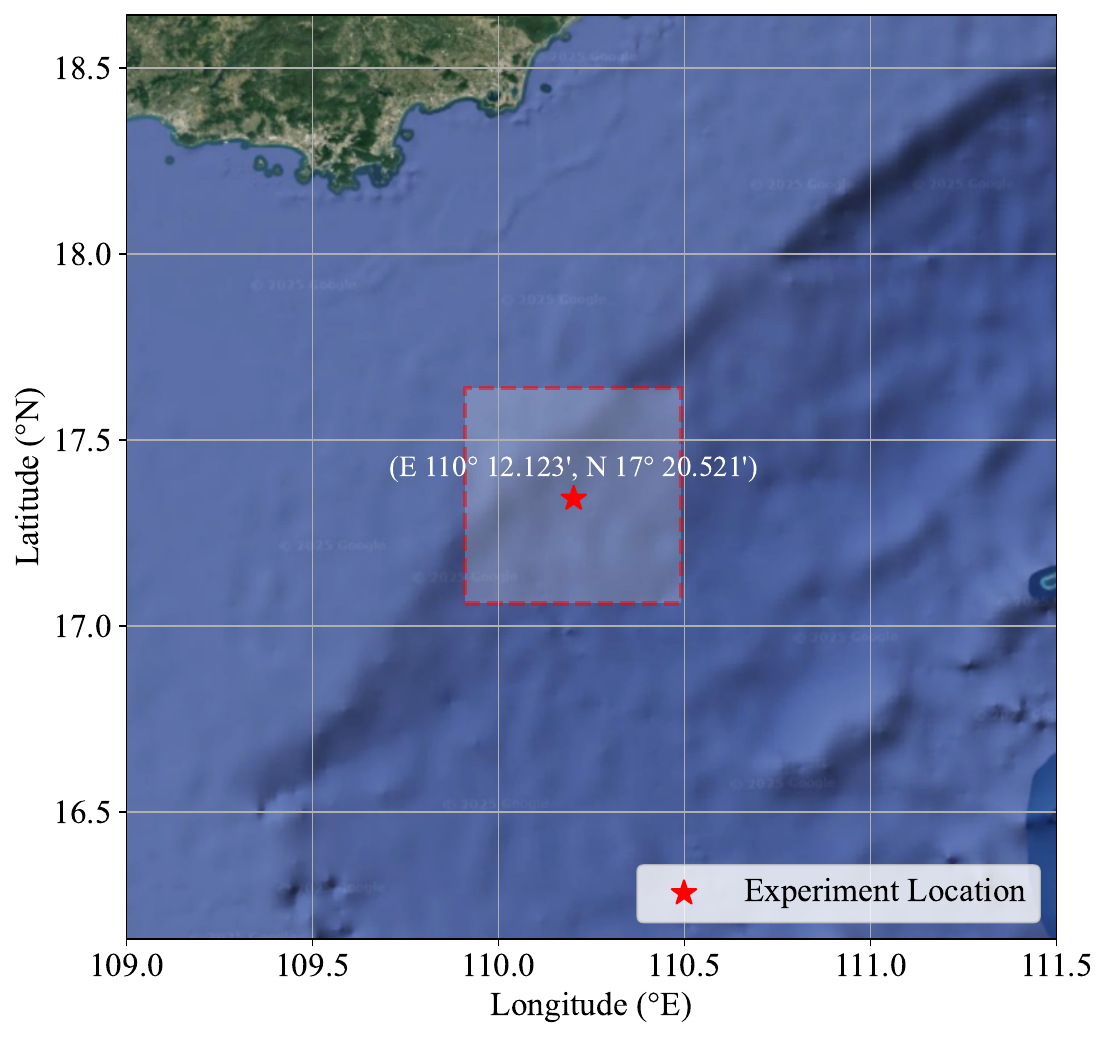}
    \label{fig:field_exp_google}
  }
  \hspace{0.05\textwidth}
  \subfloat[]{
    \includegraphics[width=0.8019\columnwidth]{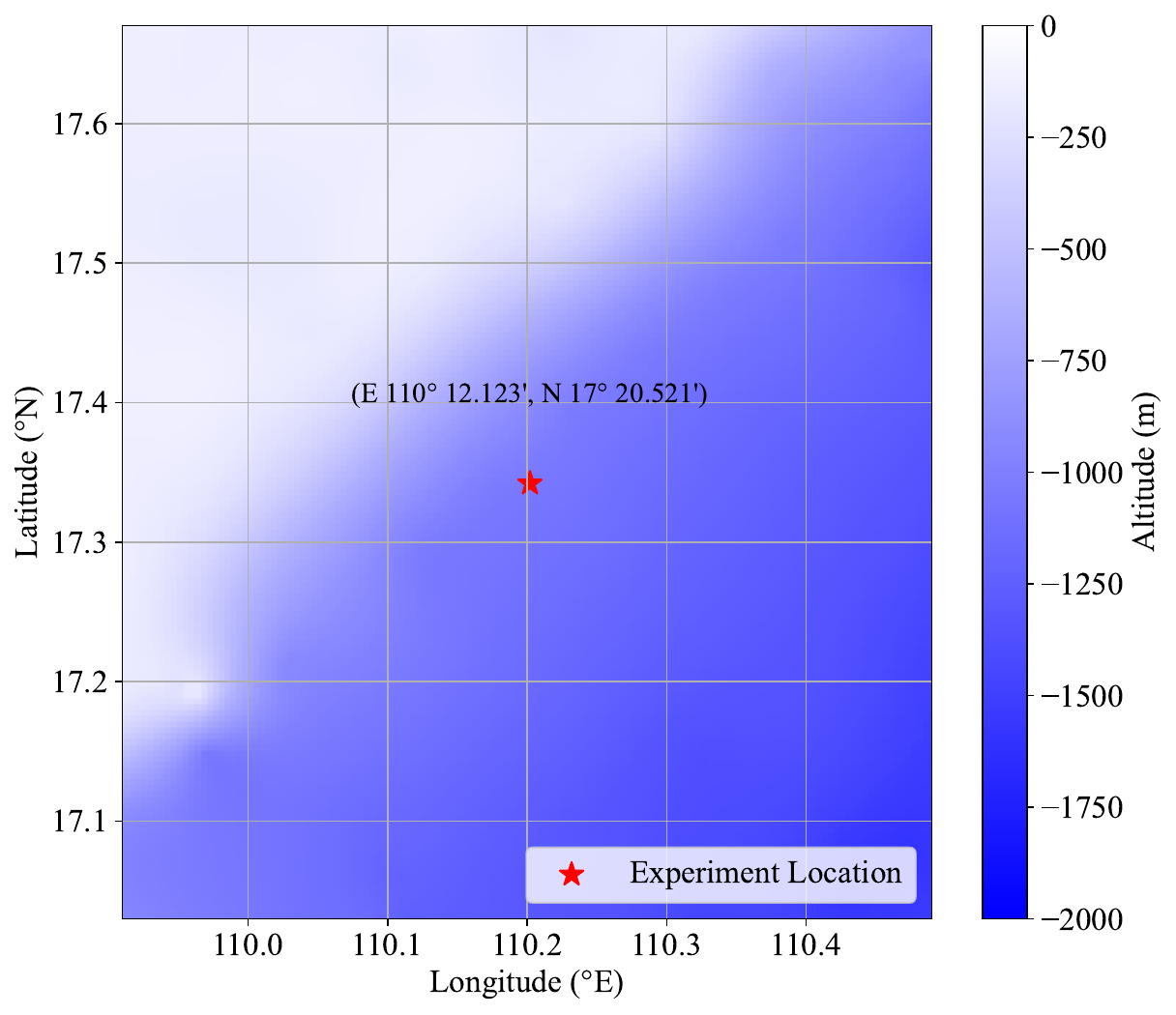}
    \label{fig:field_exp_deep}
  } 
  \caption{Overview of the field experiment site. (a) Field Experiment Location (Google Maps view). (b) Bathymetry of the Field Experiment Area.}
  \label{fig:field_exp_loc_bath}
\end{figure*}

% \begin{figure*}[!h]
%   \centering
%   \subfloat[]{
%     \includegraphics[width=0.75\columnwidth]{figs/map/prototype_device.png}
%     \label{fig:prototype_device}
%   }
%   \hspace{0.05\textwidth}
%   \subfloat[]{
%     \includegraphics[width=0.715\columnwidth]{figs/map/field.pdf}
%     \label{fig:field_exp_on_site}
%   }
%   \caption{Prototype and On-site Views of the Field Experiment. (a) Photo of the prototype device used in the field experiment. (b) On-site photo taken during the field experiment.}
%   % \label{fig:field_exp_loc_bath}
% \end{figure*}

\begin{itemize}
  \item[$\bullet$] Except for the m-model, the $F_1$ scores of StreakNetv2 on s, l, and x models is higher than StreakNet, indicating that the average anti-scattering performance of DBC-Attention is superior to Self-Attention.
  \item[$\bullet$] The number of network parameters of s, m, l, x models increase sequentially. The $F_1$ scores of StreakNet and StreakNetv2 increases from s to m and decreases thereafter, indicating varying degrees of overfitting in both architectures after the m-model. Although StreakNet's performance is significantly higher than StreakNetv2 on the m-model, it significantly decreases on the x-model. Overall, StreakNet shows large fluctuations in anti-scattering performance from s to x, while StreakNetv2 remains relatively stable, indicating that DBC-Attention has stronger anti-overfitting performance than Self-Attention in underwater imaging tasks.
\end{itemize}

We simultaneously conducted ablation experiments with traditional imaging methods. We performed Attention Analysis on the FD Embedding Layer of both StreakNet and StreakNetv2 (Use StreakNet-Emb. and StreakNetv2-Emb. to denote them, respectively). The results were used as equivalent filters, replacing the traditional 450 MHz - 550 MHz bandpass filter for imaging. The results on the validation set are presented in Fig. \ref{fig:f1_streaknet_embedding} and Table \ref{tab_f1_streaknet_embedding}.

\begin{table}[htbp]
  \begin{center}
  \caption{$F_1$ scores (\%) for traditional imaging methods and AAM equivalent filtering imaging methods on the validation set.}
  \label{tab_f1_streaknet_embedding}
  \begin{tabular}{c|ccccc}
  \toprule
  \thead{Model} & \thead{Bandpass \\ (baseline)} & \multicolumn{2}{c}{StreakNet-Emb.} & \multicolumn{2}{c}{StreakNetv2-Emb.} \\
  \midrule
  s & 70.82 & 70.42 & -0.39 & 72.69 & +1.87  \\
  m & 70.82 & 71.18 & +0.36 & 73.05 & +2.24  \\
  l & 70.82 & 68.21 & -2.61 & 70.38 & -0.44  \\
  x & 70.82 & 68.37 & -2.45 & 68.71 & -2.11  \\
  \bottomrule
  \end{tabular}
  \end{center}
\end{table}

From the experimental results, it is evident that the overall performance of StreakNetv2-Emb is significantly better than StreakNet-Emb. This further demonstrates that the features learned by DBC-Attention exhibit stronger anti-scattering capabilities compared to Self-Attention.

\section{Field Experiment}

\begin{figure}[htbp]
  \centering
  \includegraphics[width=\columnwidth]{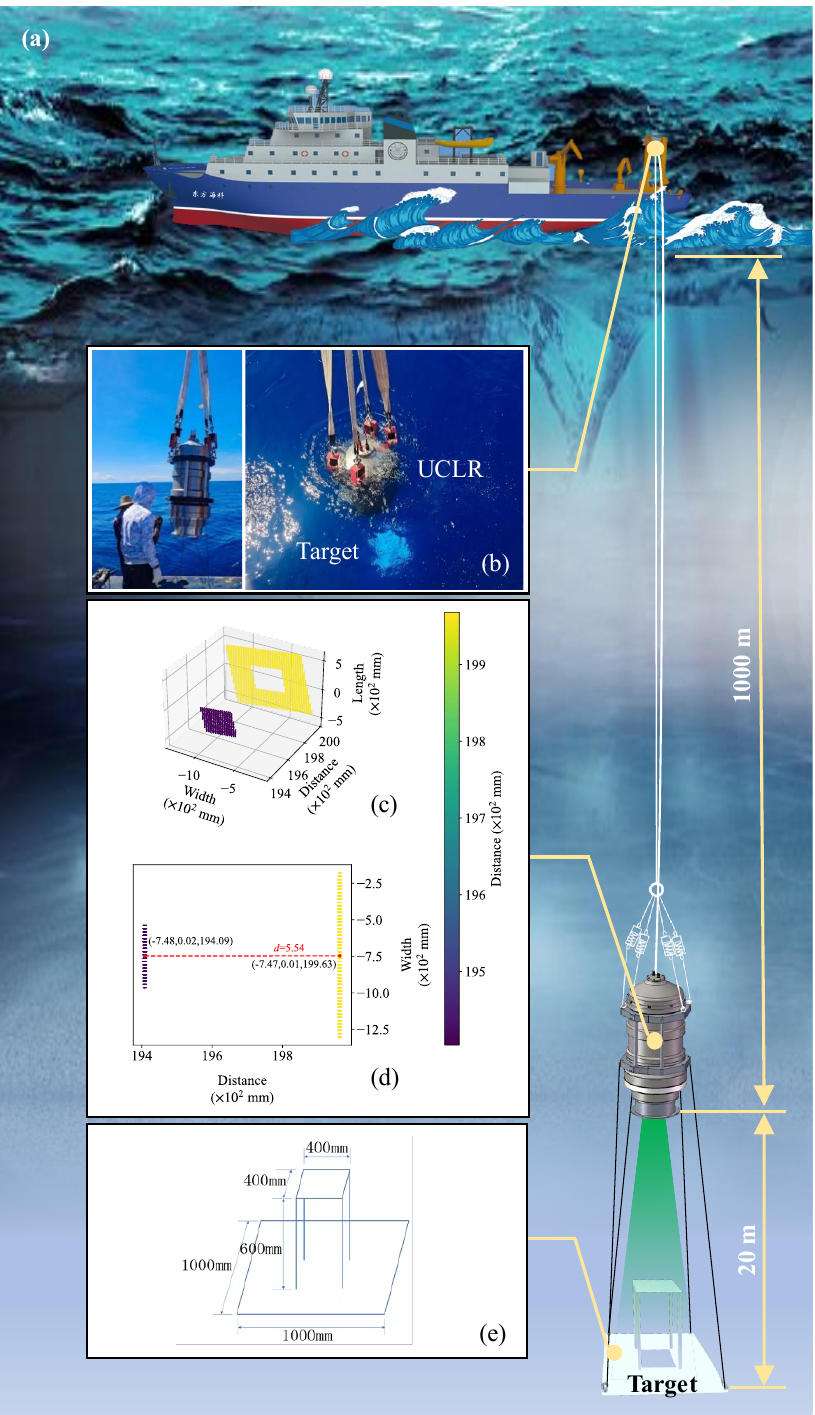}
  \caption{Setup and Results of the Field Experiment. (a) Schematic of the field experiment setup and the target object. (b) On-site photo taken during the field experiment. (c) 3D imaging results obtained from the 1000 m deep-sea experiment. (d) The measured height of the protruding platform is 554 mm, with an absolute error of 46 mm compared to the ground-truth value, corresponding to a relative error of 7.6\%. (e) Schematic of the target object.}
  \label{fig:field_exp_result}
\end{figure}

To evaluate the imaging performance of the UCLR system in deep-sea conditions, a field experiment was conducted on October 29, 2023, aboard the \textit{Dongfang Haike} research vessel in the South China Sea (E 110° 12.123', N 17° 20.521', Fig. \ref{fig:field_exp_google}). The bathymetry at the experimental site is approximately 1200 meters (Fig. \ref{fig:field_exp_deep}). During the experiment, conducted under Sea State 3 (slight seas, $\leq$ 1.25 m waves), the prototype system was deployed to a depth of 1000 m using a ship-mounted winch, and the target was suspended 20 m beneath it via an iron chain (Fig. \ref{fig:field_exp_result}a, b). As shown in Fig. \ref{fig:field_exp_result}e, the target measures 1000 mm $\times$ 1000 mm, with a 400 mm $\times$ 400 mm raised platform of 600 mm height at the center.

% \begin{figure}[htbp]
%   \centering
%   \includegraphics[width=\columnwidth]{figs/test.png}
%   \caption{Schematic of the field experiment setup and target object. The UCLR prototype was deployed 1000 m underwater, with a target object, featuring a protruding platform, suspended 20 meters below using a chain.}
%   \label{fig:field_exp_target}
% \end{figure}

Due to the challenge of manually calibrating scatter suppression for binary classification ground truth in deep-sea environments, we utilized the StreakNetv2-m model, previously trained with water tank data, to perform 3D imaging. The relative error between the measured and true height of the target protruding platform after imaging was used as the evaluation metric for imaging performance.

After imaging with the UCLR system, the measured height of the protruding platform was 554 mm (Fig. \ref{fig:field_exp_result}c, d), with an error of 46 mm compared to the true value, resulting in a relative error of 7.6\%. The results of the field experiment validate the applicability of the UCLR system in deep-sea environments.

\section{Discussion}

Although StreakNet-Arch, particularly StreakNetv2 based on DWC-Attention, demonstrates superior imaging quality in the water tank environment compared to traditional Bandpass filtering, MP models, CNNs, and Self-Attention-based StreakNet within a certain computational complexity range, the StreakNetv2 network still presents a notable risk of overfitting. A contributing factor is that the current training set consists of high-resolution 3-D point clouds that must be painstakingly hand-labeled, leaving the model dependent on fully supervised learning. For example, StreakNetv2-l and StreakNetv2-x, when reaching a computational complexity of 10 GFLOPs (Table \ref{tab_training}), achieve lower F1 scores than the smaller MP and CNN models (Table \ref{tab_f1_streaknet}). Nevertheless, those human-annotated labels enable the network to learn far richer spatial–temporal correlations than traditional algorithms can capture: supervised StreakNet-v2 not only delivers markedly higher imaging fidelity but also sustains real-time throughput, thereby retaining a decisive edge in both quality and speed. These results motivate future work on unsupervised or self-supervised formulations that can alleviate the annotation burden while preserving, or even enhancing these performance gains.

\section{Conclusion}

% This paper tackles two significant challenges impeding conventional underwater imaging algorithms: limited anti-scattering capabilities and the inability to achieve real-time processing.  We incorporated Self-Attention mechanism networks into the signal processing stage to enhance the scatter resistance capability of the UCLR. Primarily, we introduce StreakNet-Arch, a groundbreaking end-to-end binary classification architecture, to overcome the real-time imaging hurdle. Additionally, DBC-Attention, a novel variant of the Self-Attention mechanism specifically optimized for underwater environments, is proposed. Experimental results conclusively demonstrate DBC-Attention's superiority, achieving significantly better performance compared to the standard Self-Attention approach. Furthermore, we have released the first-ever public dataset containing 2,695,168 real-world underwater 3D point cloud data of streak-tube camera images. \textcolor{red}{Finally, we validated the UCLR system in a deep-sea field experiment in the South China Sea, achieving 33 mm 3D imaging accuracy at a depth of 1000 meters and a target distance of 20 meters.} This work paves the way for future research on optimizing filtering methods in underwater imaging algorithms.

This study addresses two longstanding bottlenecks in underwater imaging, pronounced susceptibility to scattering and limited real‑time throughput, by embedding self‑attention mechanisms directly into the self-developed UCLR's signal‑processing pipeline. Building on this integration, we present StreakNet‑Arch, an end‑to‑end binary‑classification framework, and DBC‑Attention, a bespoke self‑attention variant optimized for turbid aquatic scenes. Together, these innovations markedly enhance scatter resistance while sustaining real‑time performance, thereby establishing a new benchmark for high‑speed, high‑fidelity underwater imaging. 

Extensive experiments on our validation set under controlled water tank environment demonstrate that both the Self-Attention-based StreakNet and the DBC-Attention-based StreakNetv2 substantially outperform traditional bandpass filtering, and achieve higher $F_{1}$ scores than learning-based MP networks and various CNN models with comparable model sizes and computational complexity. In real-time benchmarks on an NVIDIA RTX 3060 GPU, the proposed StreakNet-Arch maintains a constant Average Imaging Time (AIT) of 54 to 84 ms regardless of the number of input frames, whereas traditional algorithms’ AIT grows linearly, from 58 ms at $N=2$ to 1,257 ms at $N=64$, confirming StreakNet-Arch’s clear advantage for large-scale, real-time imaging. 

To foster further progress, we release the first public dataset of 2,695,168 real-world underwater 3D point clouds captured by streak-tube camera. Finally, we validate the complete UCLR system in a deep-sea trial in the South China Sea, achieving an error of 46 mm at 1,000 m depth and 20 m target range. This work not only sets new benchmarks in anti-scattering performance and real-time throughput but also provides a foundation for future advances in underwater imaging filtering strategies. 

\newpage

\bibliographystyle{IEEEtran}
\bibliography{IEEEabrv, IEEEtran_StreakNet}

% \begin{thebibliography}{1}

% \bibitem{vaswani2017attention}
% A. Vaswani, N. Shazeer, N. Parmar, J. Uszkoreit, L. Jones, A. Gomez, L. Kaiser, and I. Polosukhin. ``Attention is all you need,'' \textit{Advances in neural information processing systems}, vol. 30, 2017.

% \end{thebibliography}

\begin{IEEEbiography}[{\includegraphics[width=1in,height=1.25in,clip,keepaspectratio]{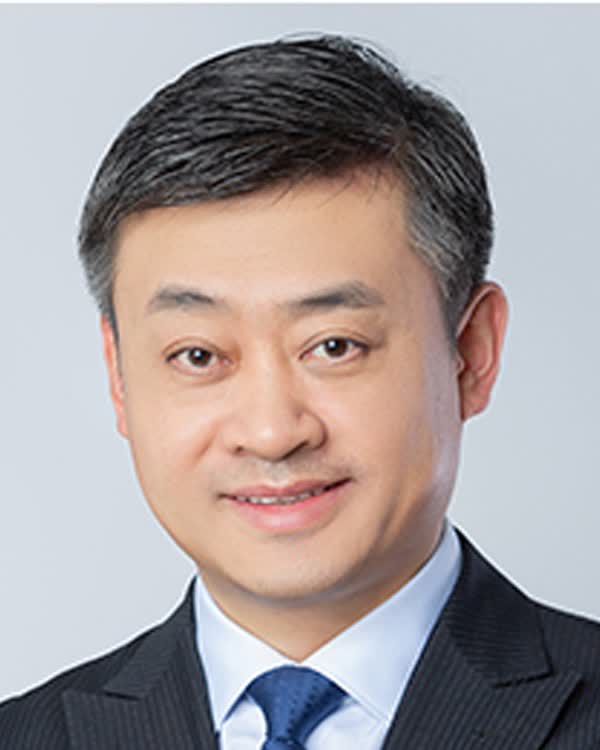}}]{Xuelong Li} is with the Institute of Artificial Intelligence (TeleAI), China Telecom, P. R. China since 2023. Before that, he was a full professor at The Northwestern Polytechnical University (2018-2023), a full professor at The Chinese Academy of Sciences (2009-2018), a Lecturer/Senior Lecturer/Reader at The University of London (2004-2009), a Lecturer at The University of Ulster (2003-2004), and he previously took positions at The Chinese University of Hong Kong, The Hong Kong University, The Microsoft Research, and The Huawei Technologies Co., Ltd.\end{IEEEbiography}

\begin{IEEEbiography}[{\includegraphics[width=1in,height=1.25in,clip,keepaspectratio]{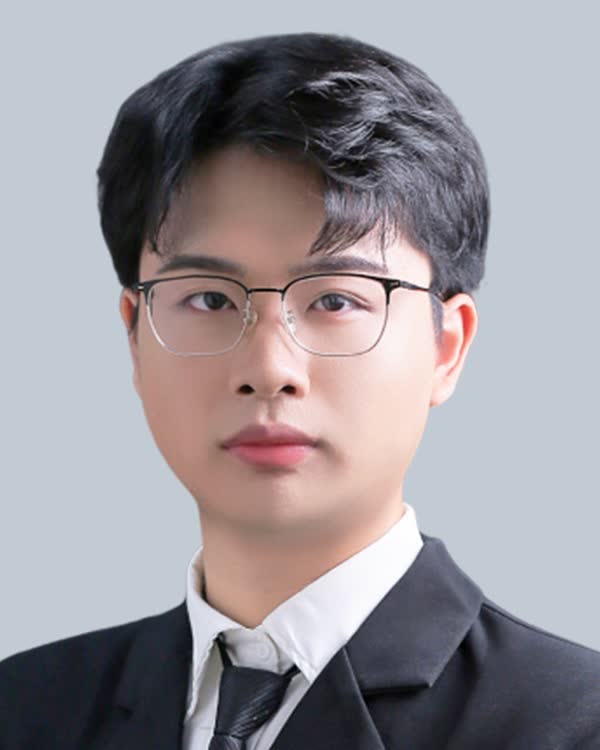}}]{Hongjun An} received the bachelor's degree in information Science and Technology College from Dalian Maritime University, Dalian, China, in 2024. He is currently pursuing the Ph.D. degree with the School of Artificial Intelligence, OPtics and ElectroNics (iOPEN) from Northwestern Polytechnical University, Xi'an, China. His research interests include water-related optics, unmanned underwater vehicles (UUVs), large models (LMs) and embodied intelligent robots. \end{IEEEbiography}

\begin{IEEEbiography}[{\includegraphics[width=1in,height=1.25in,clip,keepaspectratio]{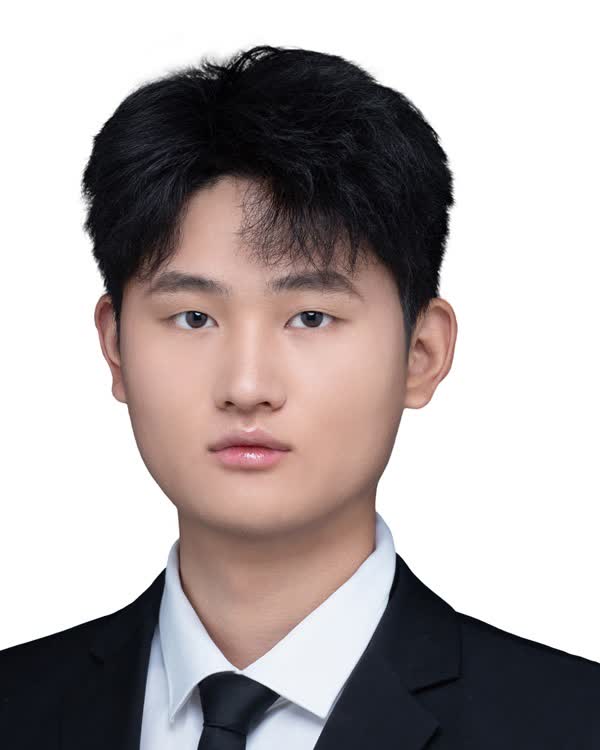}}]{Haofei Zhao} graduated with a Bachelor's degree in Information Science and Technology from Dalian Maritime University in 2024. He is currently pursuing his Master's degree in Optoelectronic Information Engineering at the School of Artificial Intelligence, OPtics and ElectroNics (iOPEN), Northwestern Polytechnical University, Xi'an, China. His research focuses on underwater optical technologies, unmanned underwater vehicles (AUVs), underwater LiDAR imaging systems, and embedded systems development for marine applications. \end{IEEEbiography}

\begin{IEEEbiography}[{\includegraphics[width=1in,height=1.25in,clip,keepaspectratio]{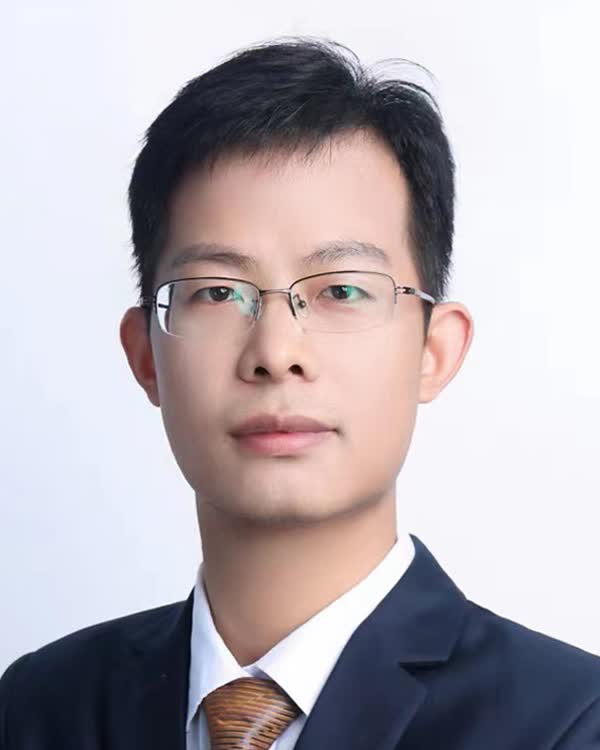}}]{Guangying Li} is the  research assistant of Xi'an Institute of Optics and Precision Mechanics, Chinese Academy of Sciences since 2022. He received his PhD degrees from University of Chinese Academy of Sciences. He is engaged in ultrafast solid-state laser technology, as well as underwater laser communication and detection technology research. \end{IEEEbiography}

\begin{IEEEbiography}[{\includegraphics[width=1in,height=1.25in,clip,keepaspectratio]{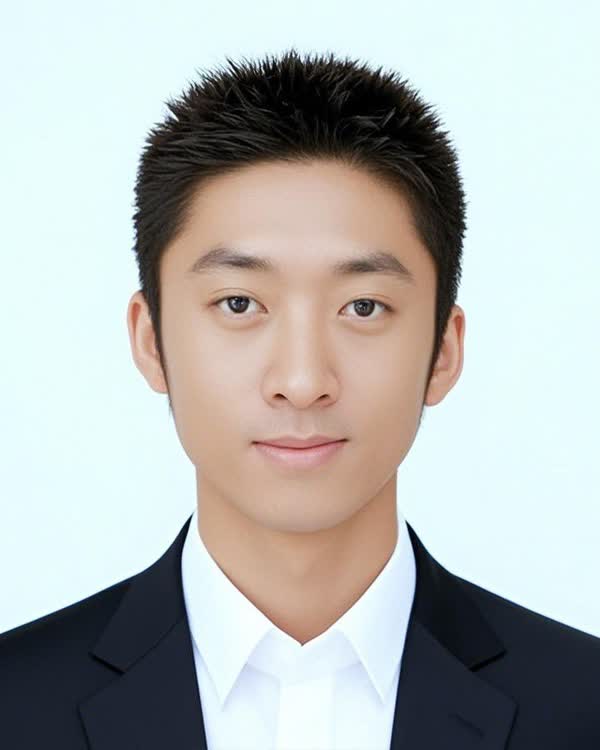}}]{Bo Liu} is now a senior engineer in the Marine Optical Technology Laboratory of Xi'an Institute of Optics and Precision Mechanics, Chinese Academy of Sciences. His research interests include optical imaging in extreme marine environments and long-range imaging. He has developed over 20 sets of underwater imaging equipment, which are widely used in China's marine scientific research and marine security fields. \end{IEEEbiography}

\begin{IEEEbiography}[{\includegraphics[width=1in,height=1.25in,clip,keepaspectratio]{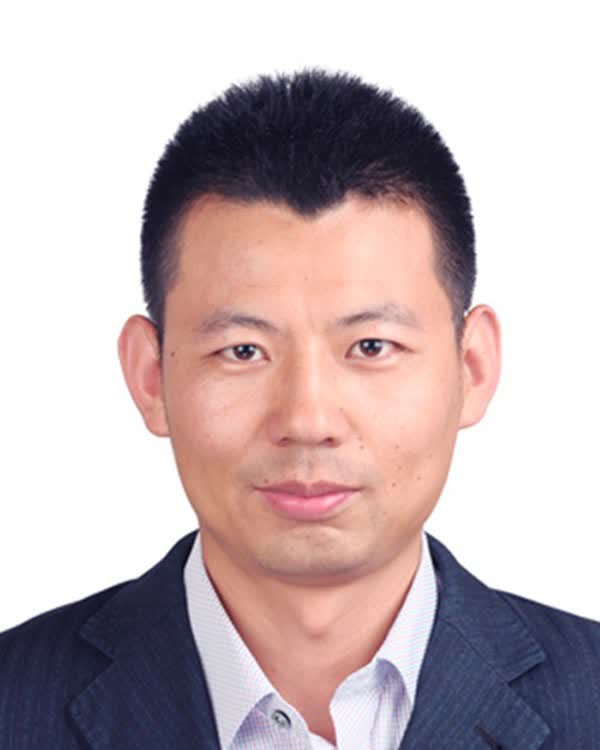}}]{Xing Wang} is currently a Professor at Xi’an Institute of Optics and Precision Mechanics, Chinese Academy of Sciences. His research interests include ultrafast and ultra-sensitive photoelectric detection devices, ultrafast diagnostic camera and 3D imaging Lidar. He has coauthored more than 50 papers. He serves as young editors of Ultrafast Sciences and Acta Photonica Sinica. \end{IEEEbiography}

\begin{IEEEbiography}[{\includegraphics[width=1in,height=1.25in,clip,keepaspectratio]{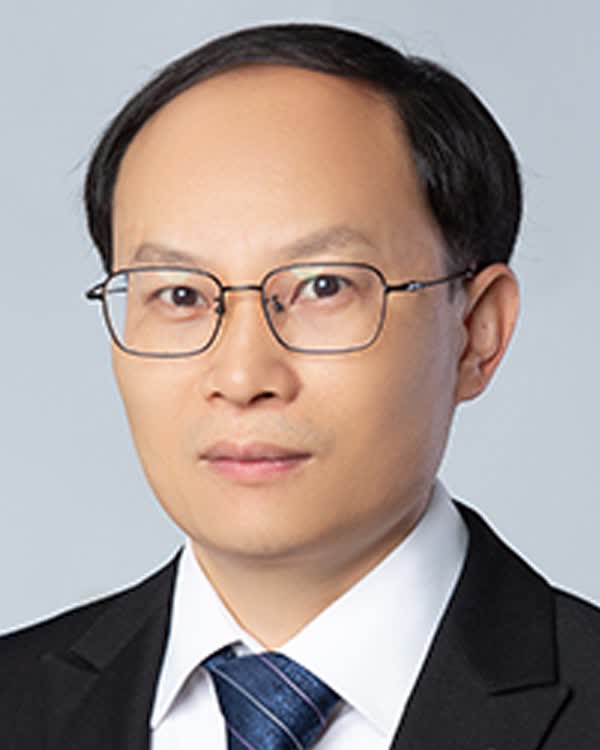}}]{Guanghua Cheng} is currently a Professor in School of Artificial Intelligence,Optics and Electronics(iOPEN), Northwestern Polytechnical University, Xi’an, China. Also, he is a visiting Professor in Laboratoire Hubert Curien, UMR 5516 CNRS, Université Jean Monnet, Saint Etienne, France. His research interests include Interaction between ultrafast laser and mater, ultrafast laser machining, high power solid laser technique, and nonlinear optics.  \end{IEEEbiography}

\begin{IEEEbiography}[{\includegraphics[width=1in,height=1.25in,clip,keepaspectratio]{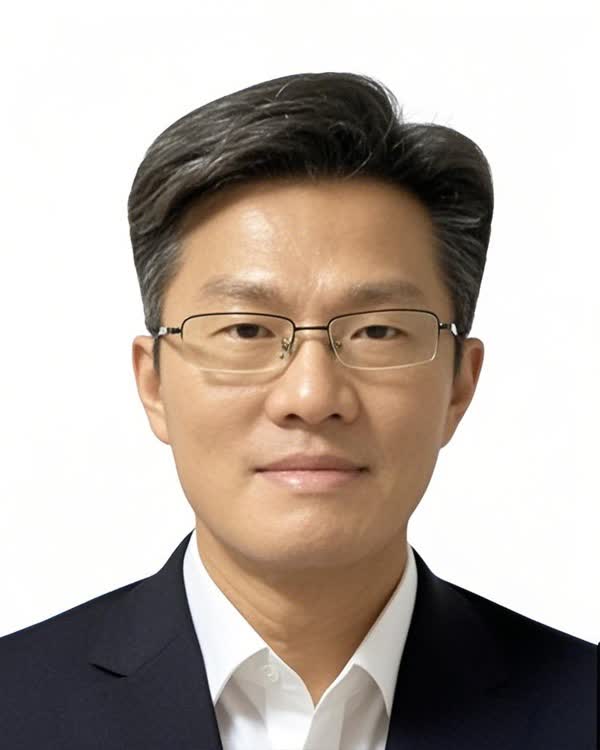}}]{Guojun Wu} is now a Professor in the Marine Optical Technology Laboratory of Xi'an Institute of Optics and Precision Mechanics, Chinese Academy of Sciences. My main research areas are ocean optical sensing technology and flow field optical measurement technology. We have successively organized and completed the research and development of deep-sea high-definition camera, wet swappable optoelectronic connectors, and multiple types of in-situ sensors for marine biogeography (chlorophyll, dissolved oxygen, nitrate, etc.).  \end{IEEEbiography}

\begin{IEEEbiography}[{\includegraphics[width=1in,height=1.25in,clip,keepaspectratio]{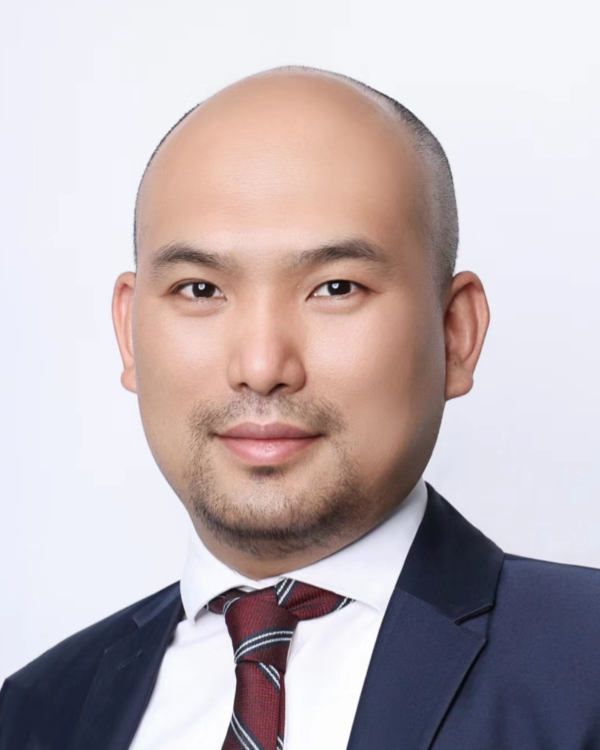}}]{Zhe Sun} is currently an Associate Professor of Northwestern Polytechnical University since 2022. Before that, he was a postdoc at Friedrich Schiller University Jena (2020-2022) and Helmholtz Institute Jena, GSI Helmholtzzentrum für Schwerionenforschung GmbH (2018-2020). Previously, he contributed as a research assistant at the Xi'an Institute of Optics and Precision Mechanics, Chinese Academy of Sciences (2014-2018). He received his PhD degrees from the Beijing University of Technology. His research interests include water-related optics, computational imaging, laser imaging. \end{IEEEbiography}

\end{document}